\documentclass{article}

\usepackage[scaled=0.85]{DejaVuSansMono}

\renewcommand{\texttt}[1]{{\fontfamily{DejaVuSansMono-TLF}\selectfont #1}}

\usepackage{minted} 
\setminted{
    fontfamily=DejaVuSansMono-TLF,
    fontsize=\small
}

\usepackage[preprint]{neurips_2026}

\usepackage[utf8]{inputenc} 
\usepackage[T1]{fontenc}    
\usepackage{url}            
\usepackage{booktabs}       
\usepackage{amsfonts}       
\usepackage{nicefrac}       
\usepackage{microtype}      
\usepackage{xcolor}         
\usepackage{lipsum}
\usepackage{pgfplotstable}

\usepackage{amsmath}
\usepackage{amssymb}
\usepackage{mathtools}
\usepackage{enumitem}

\usepackage{amsthm}

\usepackage{multirow}       

\usepackage{url}
\usepackage[utf8]{inputenc} 
\usepackage[T1]{fontenc}    
\usepackage[breaklinks]{hyperref}      

\usepackage{booktabs}       
\usepackage{amsfonts}       
\usepackage{amsmath}
\usepackage[mathscr]{euscript}
\usepackage{amssymb}
\usepackage{mathtools}
\usepackage{amsthm}
\usepackage{nicefrac}       
\usepackage{microtype}      

\usepackage{graphicx}
\usepackage{xcolor}      

\usepackage{caption}
\usepackage{multirow,makecell}

\usepackage{xspace}
\usepackage{subcaption}

\usepackage{algorithm}
\usepackage{algorithmic}
\usepackage{wrapfig}

\usepackage[most]{tcolorbox}

\usepackage[capitalize,noabbrev]{cleveref}

\title{Asynchronous Reasoning: Training-Free Interactive Thinking LLMs}

\author{%
  \!\!\!\!\!\!\!\!George Yakushev$^*$ \\ Yandex, HSE University \\ \And
  \!\!\!\!\!\!\!\!Nataliia Babina$^*$ \\ The University of Tokyo, MATS \\ \And
  \!\!\!\!\!\!\!\!Masoud Vahid Dastgerdi$^*$ \\ HSE University \\ \And
  \!\!\!\!\!\!\!\!Vyacheslav Zhdanovskiy$^*$ \\ Yandex \\ \And
  \!\!\!\!\!\!\!\!Denis Kuznedelev$^*$ \\ Yandex \\ \And
  \!\!\!\!\!\!\!\!Alina Shutova \\ Yandex, HSE University \\ \And
  \!\!\!\!\!\!\!\!Max Ryabinin$^\dagger$ \\ Together AI \\ 
}
\begin{document}
\maketitle
\begin{abstract}
Many state-of-the-art LLMs are trained to think before giving their answer.
Reasoning can greatly improve language model capabilities, but it also makes them less interactive: given a new input, a model must stop thinking before it can respond.
Real-world use cases such as voice-based or embodied assistants require an LLM agent to respond and adapt to additional information in real time, which is incompatible with sequential interactions.
In contrast, humans can listen, think, and act asynchronously: we begin thinking about the problem while reading it and continue thinking while formulating the answer. 
In this work, we augment LLMs capable of reasoning to operate in a similar way without additional training. 
Our method uses the properties of positional embeddings to enable LLMs built for sequential generation to simultaneously think, listen, and write outputs.
We evaluate our approach on math, commonsense, and safety reasoning: it allows models to generate accurate thinking-augmented answers while reducing time to first non-thinking token from minutes to ${\le}$ 5s and the overall delays by up to $12{\times}$.
\end{abstract}

\vspace{-20px}
\section{Introduction}\label{sect:intro}
\vspace{-4px}

Modern language models solve complex tasks using inference-time computation mechanisms~\cite{scaling_test_time_snell2024scaling,challenging_bigbench_solved_with_cot_Suzgun2022ChallengingBT,beeching2024scalingtesttimecompute}, such as chain-of-thought reasoning~\cite{cot_wei_2022,zero_shot_cot_Kojima2022LargeLM,tree_of_thought,verify_step_by_step}\nocite{auto_cot_Zhang2022AutomaticCO,muennighoff2025s1} and agentic tool use~\cite{Schick2023ToolformerLM,Yao2022ReActSR, pmlr-v202-gao23f}\nocite{Shen2023HuggingGPTSA,Qin2023ToolLLMFL,azerbayev2024llemma, wang2024mathcoder, li2024chainofcode}.
Recent models, both proprietary~\cite{openai_o1,googledeepmind2025gemini25thinking,AnthropicClaude3.7Sonnet} and open~\cite{deepseek_r1,qwq32b,kimik2openagentic}, are explicitly trained for reasoning and agentic capabilities.
As we trust these models with harder problems~\cite{openai_arc_prize_o3,HumanitysLastExam2025}, their ability to ``think'' becomes ever more important.
%
The current dominant strategy for large language model (LLM) reasoning is the \textit{read-think-answer} cycle: the model encodes a given problem, generates chain-of-thought reasoning, possibly calls tools, and then formulates the final answer~\cite{openai_o1,deepseek_r1,kimik2openagentic}. 
This paradigm naturally fits the sequential view of LLMs as next token prediction models.
However, it also means that the LLM must follow a rigid turn structure that can limit its flexibility. 
The ``thinking'' phase can take minutes of real time, during which the agent does not get new information or output its current results.

By contrast, people have an innate ability to think asynchronously~\cite{human_reading_brainscans,human_Lyu2023Finding,human_Trasmundi2023Mind,human_jintelligence13120156}. 
We can begin solving a problem even before hearing its entire statement. 
Moreover, we might start talking (or acting) while still completing our solution.
Such ``multitasking'' is not always easy or efficient~\cite{human_Madore2019Multicosts}, but it allows us to operate effectively in a dynamic environment~\cite{corbetta2008reorienting}.

Similarly, artificial agents often need the ability to dynamically change their course of action.
A voice assistant is expected to maintain a conversation in real time~\cite{audio_gpt4o, audio_palm, audio_speechgpt, audio_qwen, mini_omni, llama_omni, audio_moshi}. 
The VLA model of an embodied agent needs to quickly adjust to new inputs~\cite{vla_Ahn2022SayCan, vla_Brohan2023RT2, vla_Driess2023PaLME}.
Even fully text-based ``deep research'' agents benefit from interactive communication with the user~\cite{openai2025deepresearchinteractivity}.
However, the standard read-think-answer cycle is inherently non-interactive. 
During the thinking phase, if an agent receives new inputs or must take action, it can either stop reasoning, discarding any incomplete thoughts, or wait until the thinking phase is complete, sacrificing interactivity. 
As a result, many real-time LLM applications do not fully benefit from thinking.

\begin{figure*}[t]
    \centering
    \vspace{-35px}
    \includegraphics[width=\linewidth]{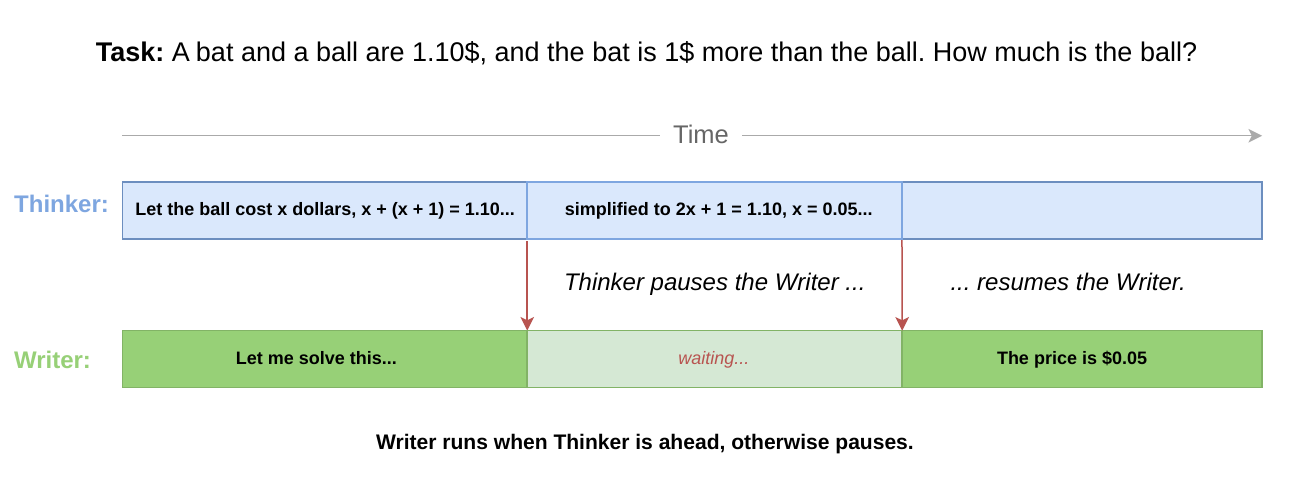}
    \vspace{-20pt}
    \caption{An intuitive explanation of asynchronous reasoning: the model generates its response concurrently with thinking. If the thinking stream needs additional time, it can pause the writing stream until the next reasoning step is ready. See animated version in the supplementary code.}
    \label{fig:teaser}
    \vspace{-15pt}
\end{figure*}

In this work, we propose a technique that enables asynchronous LLM reasoning.
Instead of retraining the model to satisfy each specific degree of interactivity, we propose an approach that leverages existing models, only changing their operation at inference time.
Our approach uses three concurrent token streams: user inputs, private thoughts, and a public response, all of which can be updated in real time.
We rely on the geometric properties of rotary positional embeddings to make the LLM perceive these streams as a single contiguous sequence \textit{without additional training}.
The model itself can decide whether it should continue talking or pause and think, depending on the current state of the three streams. 
The resulting asynchronous reasoning method can be formulated as standard LLM inference with a modified attention cache, making it possible to easily integrate our approach into efficient LLM inference frameworks~\cite{kwon2023efficient,zheng2024efficiently}.
Our main contributions can be summarized as follows:
\begin{itemize}[leftmargin=*]
    \vspace{-4px}\item We propose AsyncReasoning, a zero-shot method that allows existing reasoning LLMs to think, write outputs, and encode extra inputs concurrently. 
    Our approach relies on model-agnostic concurrent attention and zero-shot mode switching, making it easy to adapt to new models.
    \vspace{-4px}\item We evaluate AsyncReasoning on multiple real-time benchmarks in mathematical and commonsense reasoning. 
    Our experiments demonstrate that the proposed approach lets the LLM overlap thinking and answering, reducing time to first token by up to $80\times$ and total user-perceived delay by $12\times$ while retaining accuracy gains from reasoning. 
    \vspace{-4px}\item We demonstrate how AsyncReasoning can be used to improve model safety by thinking about risks in the background.
    This allows streaming real-time outputs on benign requests, while 
    considering the safety implications in a private thinking stream that can pause potentially harmful outputs.
    
    \vspace{-4px}\item Finally, we evaluate on tasks where the user specifies additional information after the initial prompt, such as clarifications or error correction. We found that modern LLMs with AsyncReasoning can incorporate this information on the fly without interrupting reasoning.

    \vspace{-4px}\item We provide a reference implementation of AsyncReasoning\footnote{See \href{https://github.com/yandex-research/AsyncReasoning}{\texttt{github.com/yandex-research/AsyncReasoning}}}\!, including GPU kernels for concurrent attention and a minimal voice assistant with asynchronous thinking capabilities.
\end{itemize}

\vspace{-10px}
\section{Related Work}\label{sect:background}
\vspace{-8px}
\subsection{Real-time LLM Applications}
\vspace{-5px}
Modern LLM agents are deployed in a wide range of applications that require varying degrees of interactivity.
Even with text-based AI assistants, users often interrupt inference to provide missing context or corrections~\cite{anthropic2026agents}, but they do not always expect quick responses.
For instance, a background code review agent can pause and think for several minutes, whereas a real-time voice assistant cannot.
Here, we briefly review several LLM applications that require quick or interactive responses.

\textbf{Voice assistants.} Recent works~\cite{audio_palm,audio_speechgpt} and industry releases~\cite{audio_gpt4o,gemini_live,claude_voice_mode_2025} use LLM agents as interactive voice assistants that talk to users in real time, often through phones or edge devices, or participate in a group conversation~\cite{Flamino2025_testing_the_limits,agent_brainstorming}.
Compared to their text-based counterparts, voice assistants need faster reaction time, with the user often adding information while the agent is thinking.

There are two main strategies for building voice assistants: modular and end-to-end. 
The first strategy feeds the output of automated speech recognition~\cite{first_asr_davis1952,kaldi,wav2vec,whisper} into a text-based LLM, then sends its response into a text-to-speech (TTS) system~\cite{first_tts_umeda1968,ssps2009,wavenet,tacotron,tacotron2,waveglow,hifigan,tortoisetts}. 
The pipeline overlaps the LLM generation with TTS to stream audio in real time.
The second, more recent, strategy is using language models that are trained to process and generate audio natively, often called omnimodal or speech language models
~\cite{audio_qwen,audio_moshi,mini_omni,llama_omni}. 
However, due to the constraints on response time, many speech LMs are not trained for long-form reasoning, and thinking models often cannot generate speech. 
For example, in the recent Qwen3-Omni family~\cite{Qwen3-Omni}, the \href{https://huggingface.co/Qwen/Qwen3-Omni-30B-A3B-Instruct/blob/26291f793822fb6be9555850f06dfe95f2d7e695/README.md}{30B-A3B-Instruct} model can speak, but does not generate \texttt{<think>} blocks, while the \href{https://huggingface.co/Qwen/Qwen3-Omni-30B-A3B-Thinking/blob/2f443cfc4c54b14a815c0e2bb9a9d6cbcd9a748b/README.md}{30B-A3B-Thinking} model has no speech \textit{output}.\nocite{salmonomni1,salmonomni2,omniflatten}


\textbf{Robotic \& virtual agents.} 
Another type of LLM applications that requires interactivity is agents in real-time environments. 
Agents controlling robotic systems use Embodied Language Models~\cite{vla_Driess2023PaLME,vla_Ahn2022SayCan,vla_Mon-Williams2025ELLMER,vla_Wang2023Voyager,vla_Jiang2023VIMA} for action planning or Vision-Language-Action~\cite{vla_Brohan2023RT2, vla_openVLA, vla_survey} models to control the system directly. 
Aside from robotic systems, similar agents have been proposed for videogames~\cite{llm_minecraft} or managing operating systems and mobile devices~\cite{llm_manage_linux,llm_oscopilot,zhang2023appagentmultimodalagentssmartphone,llm_autonomous_attack}. 
Similar to voice assistants, embodied agents need to react quickly to new stimuli from the environment.

\vspace{-4px}
\textbf{Reasoning and Safety.} Multiple works have studied the interactions between LLM reasoning and the safety of model's responses~\cite{cot_monitorability_fragile,Baker2025MonitoringRM}. 
By default, thinking can both mitigate safety risks and create new ones~\cite{safety_risk_and_boon_dual,safety_risk_chua2025thoughtcrimebackdoorsemergent,safety_risk_wang2025cost}. 
However, when specifically prompted to reason about the safety implications of their task, language models can detect and prevent jailbreak attacks~\cite{zhou2025safekeyamplifyingahamomentinsights,safety_boon-lou-etal-2025-think,safety_thinking_intervention,zhang2025safereasoninglargereasoning}. We review these works in Appendix~\ref{app:safety_background}. 
At the same time, standard reasoning delays the model's response, which is inconvenient for interactive usage. 
We show that LLMs can reason about the safety of the user's request in the background, mitigating jailbreaks without response delays. 

\vspace{-10pt}
\subsection{Efficient LLM Reasoning}
\vspace{-7pt}

As we have shown above, there is a wide range of tasks that require LLMs to reason in real time. 
However, most thinking models~\cite{openai_o1,deepseek_r1,qwen3} follow a read-think-answer cycle, which is inherently non-interactive. 
When receiving new information mid-thought, such LLMs can either interrupt their reasoning to react, sacrificing any incomplete thought tokens, or continue reasoning non-interactively.

Recently, there has been a large influx of techniques for efficient reasoning~\cite{efficient_reasoning_survey} through more concise chain-of-thought~\cite{efficient_reasoning_cod,efficient_reasoning_sot,efficient_reasoning_thinkless}, adaptive reasoning effort~\cite{efficient_reasoning_thinkswitcher,efficient_reasoning_switchcot,efficient_reasoning_fastthinking} or early stopping~\cite{efficient_reasoning_thoughtterminator,efficient_reasoning_stopwhenenough,efficient_reasoning_laaouach2025haltcot}. 
Another line of work explores reasoning in parallel, with multiple concurrent LLM instances solving different subtasks~\cite{ning2024skeletonofthought,jin2025learningpromisescalinglanguage,hogwild_inference,yu2025accelerateparallelizablereasoningparallel,groupthink,zheng2025parallelr1}, or parallel tool calling~\citep{gim2024asynchronousllmfunctioncalling,kim2024llmcompiler}. 
In this work, we focus on an orthogonal direction: instead of generally faster thinking, we let the LLM formulate its response concurrently with its reasoning to reduce delays.

\textbf{Reducing reasoning-induced delays.} 
Several recent studies propose techniques that reduce the reasoning delays for real-time applications with partial read/think overlapping~\cite{tong2025streamingthinkerlargelanguagemodels}\nocite{guo2025largelanguagemodelsreadwrite} or specialized two-model architectures with fast interactive and slow reasoning modules~\cite{wu2025mindpacedspeakingdualbrainapproach}.
A concurrent work~\cite{liang2025plantainplananswerinterleavedreasoning} introduced a method that fine-tunes reasoning LLMs to solve their task with \textit{interleaved} thinking and talking sub-blocks, making these models more interactive.
Note that all these techniques require specialized fine-tuning or training from scratch, which complicates their adoption. 
In practice, the requirements for interactive LLM use also vary with hardware and software configuration: a model trained for ``real-time'' reasoning on a B200 GPU may have delays on slower GPUs or with batched inference. 
Hence, models that were trained for one interactive use setup may need retraining for other setups. 
Instead, we design an asynchronous reasoning method that does not require training.

\vspace{-14px}
\section{AsyncReasoning}\label{sect:method}
\vspace{-10pt}

To convert a reasoning LLM into an asynchronous thinker, we need to formulate concurrent reasoning in a way that is compatible with the prompting structure the model was trained with. 
In Section~\ref{sect:method_thinker_talker_view}, we describe how to dynamically rearrange the model's context so that it views two asynchronous streams as one sequence. 
In Section~\ref{sect:method_mode_switching}, we discuss mode switching: allowing the LLM to alternate between concurrent writing and waiting for thoughts to finish. 
Finally, we discuss efficient parallel token processing and other implementation details in Section~\ref{sect:method_details}.

\vspace{-10pt}
\subsection{Dual Thinker \& Writer Views}\label{sect:method_thinker_talker_view}
\vspace{-7pt}

\begin{figure*}[t]
    \centering
    \vspace{-25pt}
    \includegraphics[width=\linewidth]{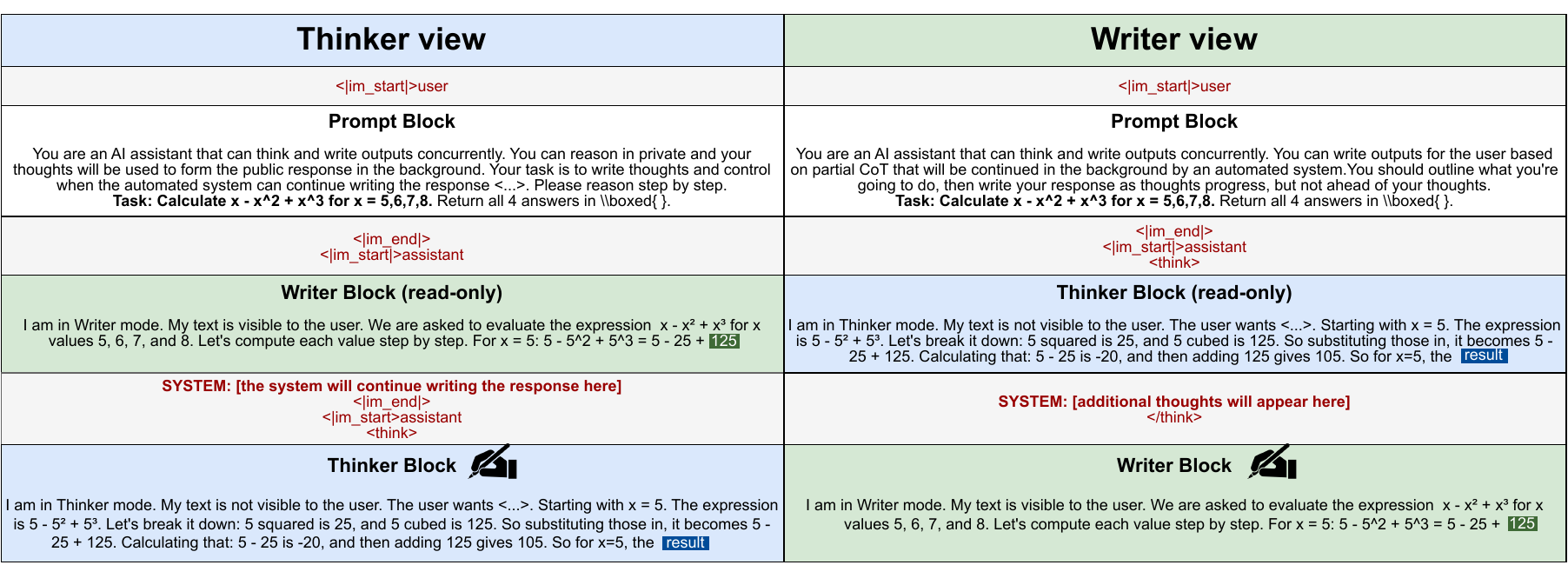}
    \vspace{-14pt}
    \caption{A dual thinker / writer view of the same reasoning task. The two views reuse the same KV cache and generate tokens in parallel. Both thinker and writer views match the model's chat template.}
    \label{fig:writer_and_thinker_view}
    \vspace{-10pt}
\end{figure*}

The core idea behind our approach is that transformer LLMs are designed to manipulate sets~\cite{vaswani2017attention,lee2019set}, and the only thing that makes them aware of \textit{sequence order} is positional encoding~\cite{relative_pos_emb,press2022trainshorttestlong,su2021roformer}. 
To change the order of elements in the prefix, we do not need to physically rearrange tokens in memory. 
Instead, it is sufficient to change positional relations between tokens, since the rest of the transformer architecture is already position-invariant.

At each inference step, AsyncReasoning manipulates positional representations to rearrange past tokens into a different order for thinking and for writing the response. 
Public response tokens ``see'' the (partial) private thoughts as if they were generated in a standard read-think-answer cycle. 
In turn, tokens within the \texttt{<think>} block ``see'' the response as if it was generated during the previous conversation turn. 
We illustrate this dual view in Figure~\ref{fig:writer_and_thinker_view} and provide prompts in Appendix~\ref{app:prompting}.

This approach allows both ``streams'' (thinking and response) to immediately attend to each other's tokens as they are generated. 
The response tokens can ``see'' and use the latest thoughts without synchronization delays. 
Likewise, the thinking ``stream'' sees the current response tokens and can pause the model's output if longer thinking is necessary. 
This also allows our implementation to encode each generated token exactly once and rearrange tokens using the geometry of relative positional embeddings (see Section~\ref{sect:method_details}).

\vspace{-8pt}
\subsection{Mode Switching}\label{sect:method_mode_switching}
\vspace{-5pt}

Another important challenge of asynchronous thinking is deciding when to synchronize. Depending on the task at hand, the thinking stream may encounter a subtask that needs more ``thinking time'' to complete. 
If this is the case, the agent should briefly pause writing the response\footnote{For voice assistants, it may be better to communicate ``Hmm, let me think about it...'', but we do not do that.} and wait for the chain of thought to progress. 
AsyncReasoning lets the model itself control synchronization.

To achieve this, we periodically ask the model if its private thoughts are still ahead of the public response, or if it should pause and think more. 
Specifically, we insert a special prompt\footnote{
{\texttt{"\dots\textbackslash n\textbackslash nWait,\! are\! my\!\! current\! thoughts\! enough\! to\!\! write\! the\! next\! paragraph\! or\! formula?\!\! (yes/no):\!\! "}}} into the thinking stream and compare the probabilities of ``yes'' vs. ``no'' as the next token. 
If the ``yes'' token is more likely, we keep thinking asynchronously. 
Otherwise, we pause the response stream until the model outputs ``yes'' again.

In our current implementation, we insert this question at the end of every paragraph or after every $T{=}20$ thinking tokens, whichever comes first. 
Crucially, once the model gives its ``yes'' or ``no'' response, we remove this prompt from its view (by hiding the corresponding key-value entries) so that it does not interfere with the model's chain-of-thought. See Appendix~\ref{app:mode_switching} for more details.

We compare different mode-switching prompts in Section~\ref{sect:experiments_initial}.
Overall, we found that existing thinking LLMs can already control asynchronous reasoning, though sometimes they do make mistakes (we elaborate on that in Appendix~\ref{app:mode-switching-failure-mode}). 
It is possible to design more advanced mechanisms, such as allowing the LLM to reason about mode switching in parallel or introducing a classifier ``head'' to decide when to pause responding. 
However, we opt to keep AsyncReasoning simple and training-free, deferring further study of mode switching to future work.

\vspace{-8pt}
\subsection{Implementation Details}\label{sect:method_details}
\vspace{-5pt}

\begin{figure*}[t]
    \centering
    \vspace{-35pt}
    \includegraphics[width=\linewidth]{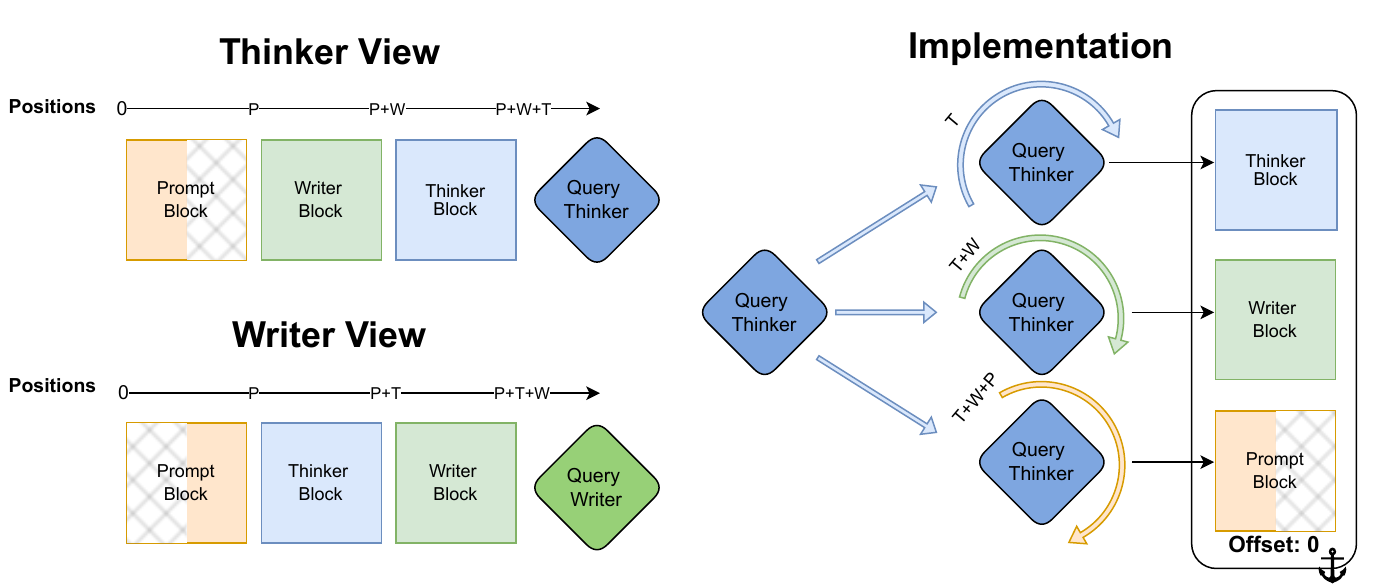}
    \vspace{-18pt}
    \caption{Concurrent thinking and writing implemented as batched inference. Newly added tokens attend to cache blocks with additional query rotations. Checkered areas represent masked tokens.}
    \label{fig:implementation_details}
    \vspace{-15pt}
\end{figure*}

In summary, AsyncReasoning arranges the thinking and response tokens in a different order depending on the generation phase, processes both streams in parallel, and periodically prompts the model to decide if it should pause and think. 
Thus, our algorithm alternates between two modes: either thinking and writing concurrently, or simply thinking while the writing is paused. 
When only one stream is active, AsyncReasoning is equivalent to standard sequential LLM inference with a combined KV cache. 
We focus the rest of this section on handling \textit{concurrent} token streams.

We implement concurrent thinking and writing by creating a custom key-value cache and adjusting the positional embeddings to account for the dual views from Figure~\ref{fig:writer_and_thinker_view}. 
The main purpose of this algorithm is to avoid redundant computation and the key-value cache bloat. 
Instead of encoding tokens twice for both views, we process each token exactly once and keep one KV cache entry that is ``viewed'' from different relative positions. 
This optimization is inspired by a similar rotation trick proposed in Hogwild! Inference~\citep{hogwild_inference}, extended to support mode switching.

\textbf{Key-Value Cache Structure.} To implement different positional views, we split the model's KV cache into three contiguous ``blocks'' (tensors): the inputs, the thinking stream, and the output stream. As new tokens are generated or added by the user, we store them in the corresponding cache block using positional representations relative to the block start\footnote{For example, given a model with RoPE embeddings, its KV cache will always store the 5th response token ``rotated'' for position 5, regardless of how many thinking tokens precede it.}.

During the self-attention forward pass, we concatenate the dot products between the query and all cache blocks, but we transform the query differently for each block to simulate the difference in token positions.
This way, the same set of attention blocks can be combined for both thinking and writing views from Figure~\ref{fig:writer_and_thinker_view} without extra memory use.

\textbf{Manipulating Positional Information.} Almost all modern LLMs use some form of relative positional information~\cite{relative_pos_emb,su2021roformer,press2022trainshorttestlong}. 
The most popular variant is rotary positional embeddings (RoPE, \citealp{su2021roformer}), which rotate query and key vectors by an angle proportional to their index in the sequence before computing the self-attention. 
Note that if both query and key are rotated by the same angle, their dot product does not change. Thus, attention only depends on the difference between query and key positions. 
In other words, rotating attention keys by $+\alpha$ is equivalent to rotating the query by $-\alpha$.

We take advantage of this property to avoid rotating the entire KV cache on each inference step. 
Instead, we keep track of the starting positions for each block and rotate the attention queries. 
Suppose there are three contiguous KV blocks: \textbf{P}rompting with $P$ tokens, \textbf{T}hinking with $T$ tokens, and \textbf{W}riting with $W$ tokens. 
When viewed contiguously (PTW), the difference between the most recent writer token and the thinker block is $T{+}W{-}1$ tokens. 
Thus, when running the forward pass for the \textit{writer}, we rotate its query by the RoPE angle corresponding to position $T{+}W{-}1$ when looking at reasoning tokens and by $W{-}1$ when looking at writer's own tokens. 
In contrast, the \textit{thinker} attends to itself at $T{-}1$ and to writer at $W{+}T{-}1$.
The same principle applies to all query-key pairs.

Formally, let $\rho(q, i)$ denote applying RoPE for vector $q$ at position $i$. The writer attends to blocks P,\! T,\! W:
$A{:=}\rho (q, i_q) {\cdot} \Bigl[ \rho (K_P, i^{P}_k), \rho (K_T, i^{T}_k), \rho (K_W, i^{W}_k) \Bigr]$, where $\left[ \cdot \right]$ denotes concatenation, $i_q$ is the query position, $i^{P}_k, i^{T}_k, i^{W}_k$ are cache block positions from the writer's point of view (see Figure~\ref{fig:implementation_details}) and $K_{P,T,W}$ are the corresponding key vectors. 
Then, we can equivalently compute attention as:
$$A {:=} \Bigl[ \rho (q, i_q {-} i^P_k) K_P,  \rho(q, i_q {-} i^T_k) K_T, \rho (q, i_q {-} i^W_k) K_W \Bigr].$$
In turn, thinker attends to the same KV cache entries with different query rotations corresponding to how they are arranged in its own view (Figure~\ref{fig:implementation_details}).
This reformulation allows us to compute $K_{P,T,W}$ once, store them in the KV cache, and only modify the attention queries for the currently processed tokens during each forward pass. Appendix~\ref{app:positional_embeddings} extends this technique to non-RoPE models.

\textbf{Technical Considerations.} 
In summary, our implementation consists of a custom KV cache and an attention kernel that uses the query rotation trick described above.
In practice, we use more than 3 KV blocks: in addition to the prompt, thinking and response tokens, we also have short linker tokens between thinking and writing blocks. 
These linkers are implemented as separate KV blocks that are visible only in one of the views (thinker or writer). 
If a block is not visible in the current view, we give it a large positional index to make it ignored due to causal attention masking.

This implementation can efficiently parallelize thinking and writing the response for small batch sizes. 
However, it can be optimized further for large batches by only processing the non-masked query-key pairs that actually contribute to the attention output. 
In future work, we plan to explore implementing more general kernels for AsyncReasoning based on vLLM's Paged Attention~\cite{kwon2023efficient}.

\vspace{-10px}
\section{Experiments}\label{sect:experiments}
\vspace{-8px}

We organize our experiments as follows: in Section~\ref{sect:experiments_initial}, we verify the design choices for each component of AsyncReasoning. 
In Section~\ref{sect:experiments_main}, we expand to additional benchmarks and models. 
Section~\ref{sect:experiments_safety} tests the security of AsyncReasoning against adversarial attacks. 
Finally, Section~\ref{sect:experiments_async_inputs} is focused on setups where the user provides additional clarifications after the initial prompt.

\vspace{-8px}
\subsection{Initial Analysis}\label{sect:experiments_initial}
\vspace{-5px}

We first evaluate different components of AsyncReasoning in detail with Qwen3-32B~\citep{qwen3}, a popular medium-sized reasoning LLM that can run on a single high-end GPU. 
We run both AsyncReasoning and baselines on one A100-SXM4 GPU in \texttt{bfloat16} precision and greedy sampling. 
We do not use general inference optimizations such as speculative decoding, as they are orthogonal to our work. 
We evaluate on MATH-500~\cite{hendrycksmath2021,verify_step_by_step}, a popular mathematical benchmark with ``medium-difficulty'' tasks that benefit from reasoning.

We focus on two main metrics: \textbf{accuracy}, computed using an LLM-as-a-judge protocol following the original benchmark setup, and \textbf{real-time delay}, defined as the amount of time (in seconds) during which the user hears no sound because the LLM has not generated the response yet.
To evaluate accuracy, we prompt the LLM to put its answer in \texttt{\textbackslash boxed\{...\}} in the public response and check its equivalence to the reference answer using the standard LLM-as-a-judge protocol\footnote{We use the evaluation protocol from \url{https://github.com/openai/simple-evals} with a \texttt{gpt-4-turbo} judge.} for MATH-500~\cite{zheng2023llm-as-a-judge}.
To measure real-time delay, we stream the assistant's response to a TTS engine (see Section~\ref{sect:method_details}) and measure the total ``silence time'': the time during which the TTS could not generate anything because the LLM is still solving the task. 
We provide a more detailed description of the TTS pipeline and a more fine-grained performance breakdown in Appendix~\ref{app:performance}.

\begin{figure}[b]
    \centering
    \vspace{-20px}

    \begin{subfigure}[t]{0.50\linewidth}
        \centering
        \includegraphics[width=\linewidth,trim=0px 20px 10px 20px,clip]{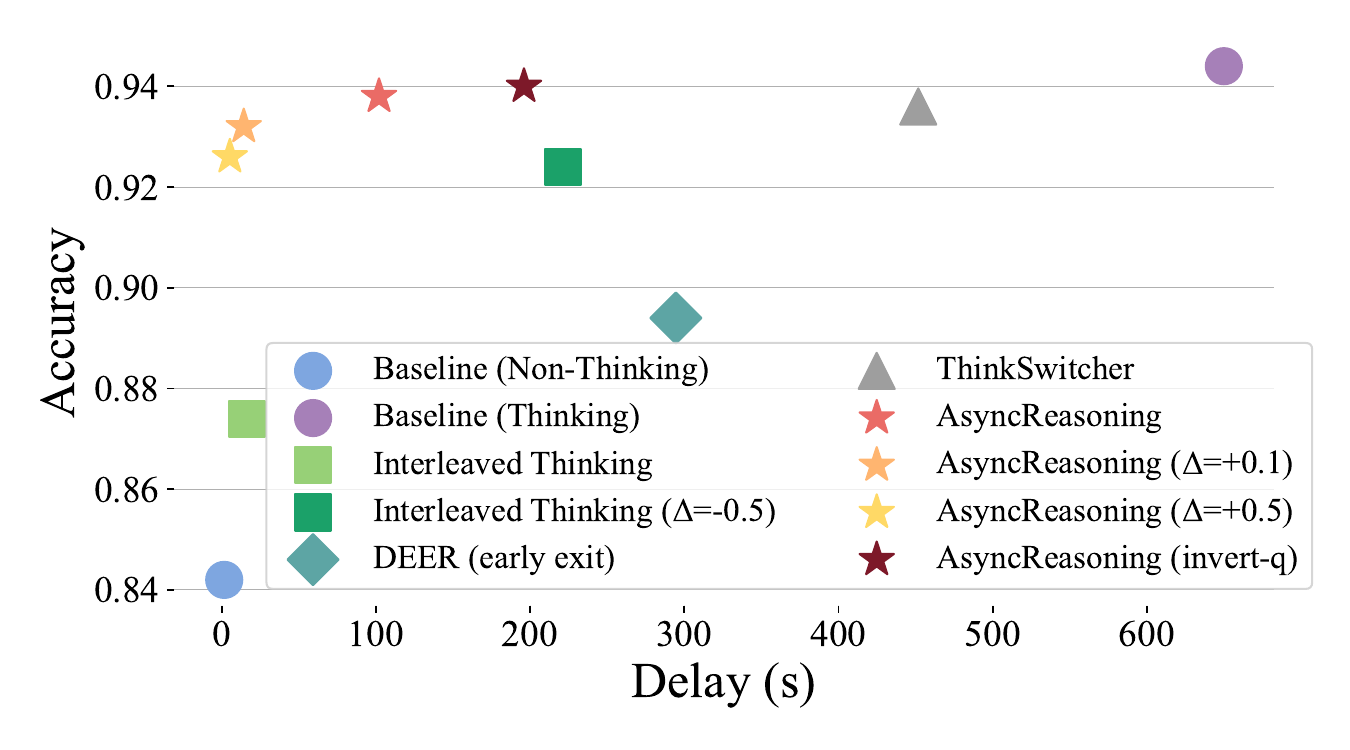}
    \end{subfigure}
    \hfill
    \begin{subfigure}[t]{0.48\linewidth}
        \centering
        \includegraphics[width=\linewidth,trim=35px 0px 50px 20px,clip]{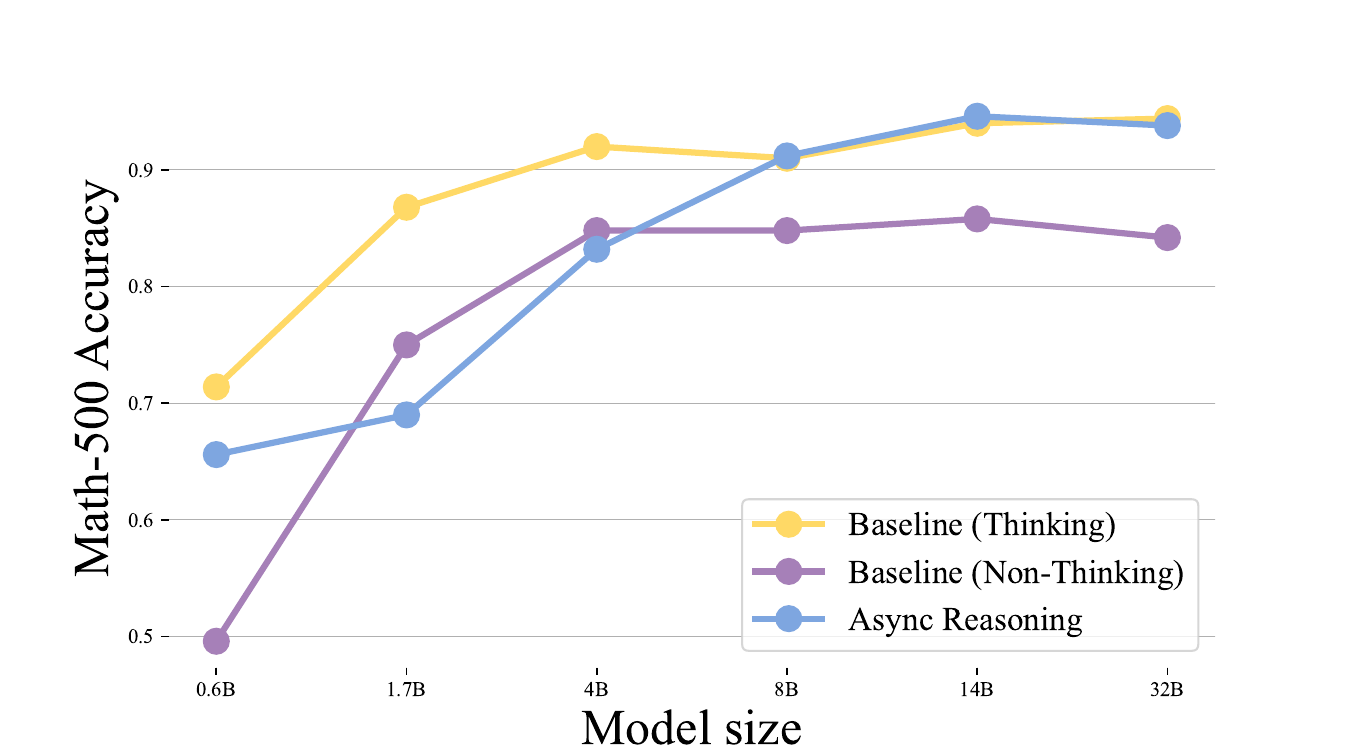}
    \end{subfigure}

    \vspace{-5px}
    \caption{MATH-500 evaluation of AsyncReasoning and baselines on A100. Left: comparison of prompts and mode switching methods on Qwen3-32B. Right: performance across Qwen3 models.}
    \label{fig:exp_math500_grouped}
    \vspace{-10px}
\end{figure}

We compare the following configurations:
\begin{enumerate}[leftmargin=*]
    \vspace{-5px}\item \textbf{Baseline (Non-thinking):} regular sequential generation with \texttt{<think>} mode disabled.
    \vspace{-3px}
    \item \textbf{Baseline (Thinking):} regular sequential generation with \texttt{<think>} mode enabled.
    \vspace{-3px}
    \item \textbf{Interleaved Thinking:} prompting the model to think and reply in short, interleaved steps, but without asynchrony. This setup is similar to Plantain~\citep{liang2025plantainplananswerinterleavedreasoning}, but without model fine-tuning.
    \vspace{-3px} \item \textbf{DEER:} training-free dynamic early exit for reasoning models, recommended parameters~\cite{yang2025dynamic}.
    \vspace{-3px} \item \textbf{ThinkSwitcher:} using a router that chooses between thinking and non-thinking mode~\cite{liang2025thinkswitcher}. We train the router on MATH (train subset) using the protocol for Qwen3-8B from the original paper. 
    \vspace{-3px}
    \item \textbf{AsyncReasoning:} the main setup from Section~\ref{sect:method_mode_switching}. The model is periodically asked whether the current thoughts are ahead of writing. If the answer is positive, the thinker and the writer both generate the next tokens in parallel. If not, the writer pauses until the next ``yes'' answer.
    \vspace{-3px}
    \item \textbf{AsyncReasoning (${+}\Delta$):} same as above, but we increase the logit for answering ``yes'' to whether the writer should continue by a fixed bias ${+}\Delta$.
    \vspace{-3px}
    \item \textbf{AsyncReasoning (invert-q):} same as above, but the question is inverted:\texttt{"\dots\textbackslash n\textbackslash nWait, should I pause writing the response and think longer? (yes/no):\!\! "}.
\end{enumerate}\vspace{-5px}

The results in Figure~\ref{fig:exp_math500_grouped}(left) show that AsyncReasoning can reduce real-time delays while preserving most of the accuracy gains from reasoning, outperforming non-asynchronous interleaved thinking. 
However, the exact tradeoff between accuracy and delay depends on the mode switching criterion and the bias. 
Our default criterion offers nearly the same accuracy as fully synchronous thinking, but at a significantly lower real-time delay. 
The ${+}\Delta$ logit bias can further reduce this delay at the cost of some accuracy loss, which comes from the writer answering too early. 
Flipping the question (where ``yes'' means pause) makes the model stop and think more often, suggesting that the model is biased to answer ``yes''. 
Unless stated otherwise, we use the main setup (no $\Delta$ or inversion) for the next sections. 
We include additional evaluations with other benchmarks, metrics (e.g., TTFT), stopping criteria and ablations on $T$ in Appendix~\ref{app:extra_ablation}.

\vspace{-4px}
\subsection{Additional Benchmarks}\label{sect:experiments_main}
\vspace{-5px}

Next, we evaluate how AsyncReasoning generalizes across problems and models. 
Initially, we targeted established speech-language benchmarks~\cite{yang2024airbench,audiobench,spoken-mqa,uro-bench}. 
However, we found that modern reasoning models can solve even the harder tasks from these benchmarks with near-perfect accuracy ($\ge95\%$) \textit{without thinking}. 
Thus, we adopt the approach from~\cite{voila2025} and use more challenging general benchmarks: MATH-500~\cite{hendrycksmath2021}, MMLU-Pro~\cite{wang2024mmlu}, GPQA-Diamond~\cite{rein2023gpqagraduatelevelgoogleproofqa}, AIME-2025~\cite{AIME2025}, ZebraLogic~\cite{lin2025zebralogicscalinglimitsllms}, and one voice-specific benchmark \texttt{spoken-mqa/multi\_step\_reasoning}~\cite{spoken-mqa}.
We report accuracy, total real-time delay and TTFT (time-to-first-token), more details in
Appendix~\ref{app:detailed_benchmarks}.

\begin{table*}[htbp]
    \centering
    \vspace{-5px}
    \caption{Evaluation of AsyncReasoning on AIME-2025
    (10 seeds) and ZebraLogic (grid) across different models using metrics from Section \ref{sect:experiments_main}. $\uparrow$ ($\downarrow$) arrows denote higher (lower) is better. Interactive baseline denotes no thinking for Qwen3 and low budget for GPT-OSS. $\dagger$---Thinking 2507}
    \small
    {\setlength{\tabcolsep}{3.7pt}
    \label{tab:exp_main_benchmarks}
    \begin{tabular}{clccccccccc}
        \toprule
        & \multirow{2}{*}{\hspace{-5px}\textbf{Model}} &
        \multicolumn{3}{c}{\textbf{Baseline (Thinking)}} &
        \multicolumn{3}{c}{\textbf{Baseline (Interactive)}} &
        \multicolumn{3}{c}{\textbf{Async Reasoning}} \\
        &  & \textbf{Acc} $\uparrow$ & \textbf{Delay} $\downarrow$ & \textbf{TTFT} $\downarrow$ & \textbf{Acc} $\uparrow$ & \textbf{Delay} $\downarrow$ & \textbf{TTFT} $\downarrow$ & \textbf{Acc} $\uparrow$ & \textbf{Delay} $\downarrow$ & \textbf{TTFT} $\downarrow$ \\ 
        \midrule

        \multirow{4}{*}{\rotatebox{90}{\textbf{AIME-2025}\;\,}}
        & \hspace{-5px}Qwen3-32B & 0.53 & 1915.2 & 1914.4 & 0.20 & 1.70 & 1.13 & 0.49 & 485.93 & 5.41 \\
        & \hspace{-5px}Qwen3-30B-A3B$^\dagger$ & 0.67 & 1814.1 & 1813.7 & 0.39 & 5.25 & 5.25 & 0.62 & 5.20 & 5.20 \\
        & \hspace{-5px}Qwen3-235B-A22B$^\dagger$\hspace{-5px} & 0.68 & 5348.2 & 5330.5 & 0.53 & 20.16 & 9.73 & 0.68 & 500.94 & 9.06 \\
        & \hspace{-5px}GPT-OSS-20B & 0.62 & 523.53 & 523.53 & 0.36 & 120.09 & 120.09 & 0.59 & 147.96 & 24.15 \\
        & \hspace{-5px}GPT-OSS-120B & 0.77 & 326.52 & 326.52 & 0.51 & 91.40 & 91.40 & 0.66 & 166.75 & 24.63 \\
        \midrule

        \multirow{3}{*}{\rotatebox{90}{\textbf{ZebraLogic}\;\,}}
        & \hspace{-5px}Qwen3-32B & 0.71 & 297.61 & 297.61 & 0.37 & 4.27 & 4.27 & 0.68 & 15.83 & 4.37 \\
        & \hspace{-5px}Qwen3-30B-A3B$^\dagger$ & 0.96 & 1244.1 & 1244.1 & 0.27 & 4.16 & 4.16 & 0.93 & 3.72 & 3.72 \\
        & \hspace{-5px}Qwen3-235B-A22B$^\dagger$\hspace{-5px} & 0.88 & 2111.7 & 2111.7 & 0.67 & 4.68 & 4.53 & 0.85 & 57.85 & 8.37 \\
        & \hspace{-5px}GPT-OSS-20B & 0.70 & 372.41 & 372.41 & 0.01 & 149.44 & 149.44 & 0.61 & 163.34 & 4.37 \\
        & \hspace{-5px}GPT-OSS-120B & 0.81 & 268.83 & 268.83 & 0.48 & 248.13 & 248.13 & 0.75 & 131.57 & 7.53 \\
        \bottomrule
    \end{tabular}
    }
    \vspace{-10px}
\end{table*}

\begin{wraptable}{r}{0.55\linewidth}
    \vspace{-12px}
    \caption{Qwen3\ SpokenMQA multistep reasoning.}
    \label{tab:exp_main_spoken_mqa}
    \centering
    {%
    \setlength{\tabcolsep}{3pt} 
    \begin{tabular}{llccc}
        \toprule
        \textbf{Size} & \textbf{Method} & \textbf{Acc} $\uparrow$ & \textbf{Delay} $\downarrow$ & \textbf{TTFT} $\downarrow$ \\ 
        \midrule
        0.6B & Baseline (Thinking) & 0.70 & 201.53 & 201.52 \\
        0.6B & Baseline (no Think) & 0.57 & 0.72 & 0.67 \\
        0.6B & AsyncReasoning & 0.66 & 2.56 & 1.37 \\
        \midrule
        4B & Baseline (Thinking) & 0.81 & 227.16 & 227.15 \\
        4B & Baseline (no think) & 0.80 & 0.80 & 0.70 \\
        4B & AsyncReasoning & 0.81 & 0.84 & 0.80 \\
        \bottomrule
    \end{tabular}
    }
    \vspace{-10px}
\end{wraptable}

\begin{figure}[!b]
    \centering
    \vspace{-15px}

    \begin{subfigure}[t]{0.49\linewidth}
        \centering
        \includegraphics[width=\linewidth,trim=0px 0px 0px 25px,clip]{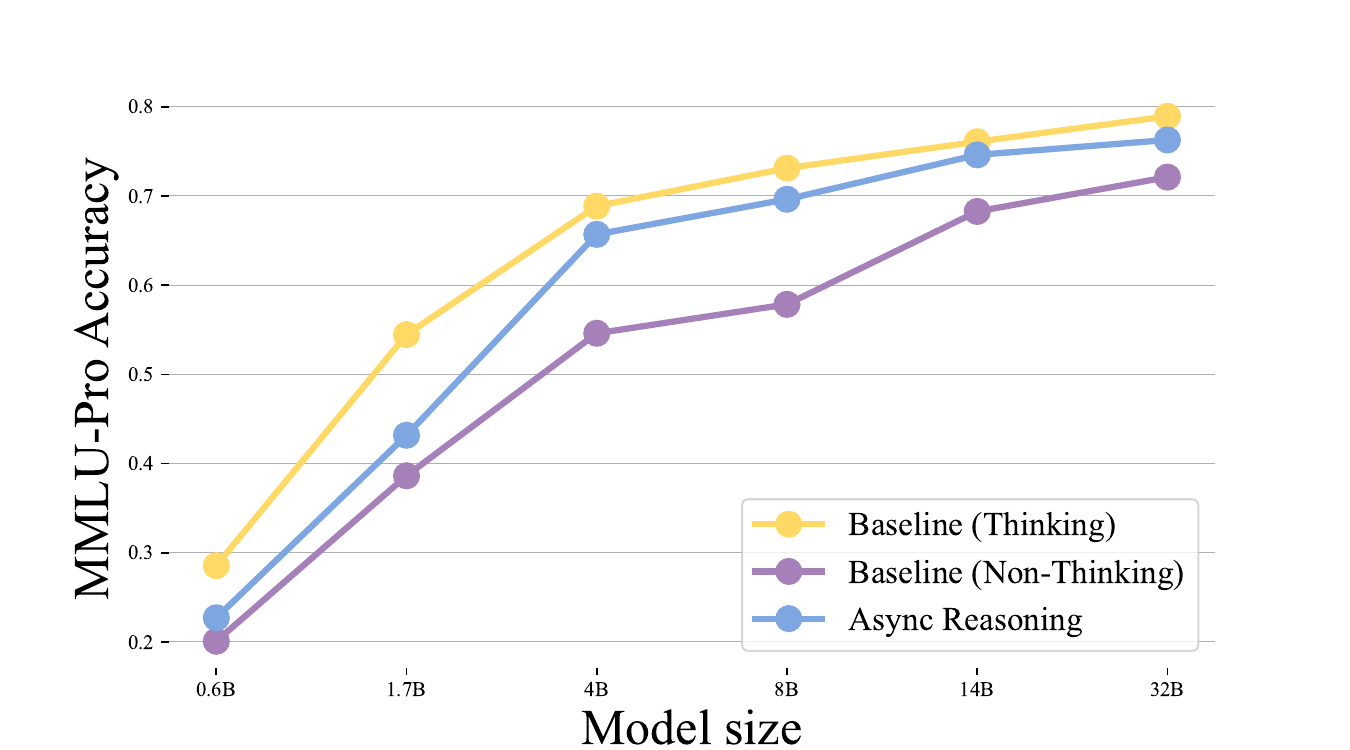}
    \end{subfigure}
    \hfill
    \begin{subfigure}[t]{0.49\linewidth}
        \centering
        \includegraphics[width=\linewidth,trim=0px 0px 0px 20px,clip]{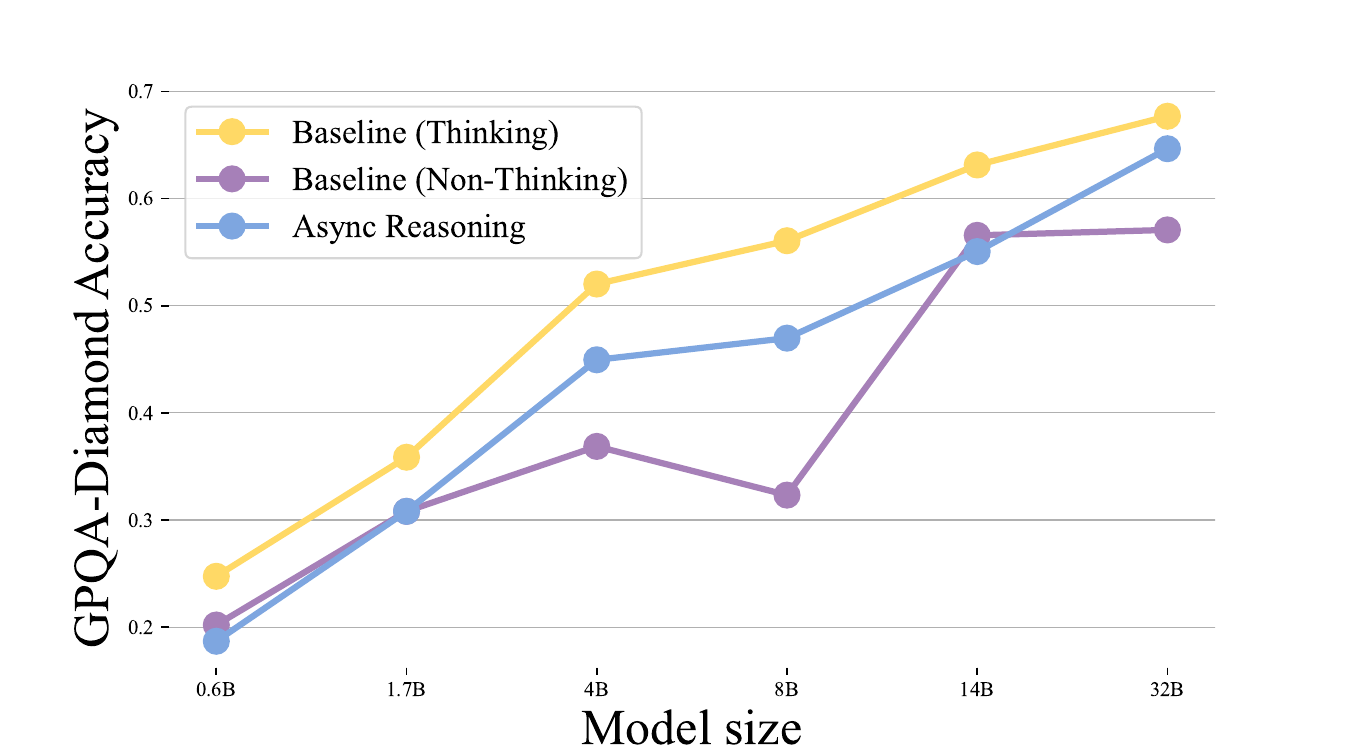}
    \end{subfigure}

    \vspace{-5px}
    \caption{Evaluating AsyncReasoning and baselines across different Qwen3 model sizes on A100 GPU for two benchmarks. Left: MMLU-Pro. Right: GPQA-Diamond.}
    \label{fig:exp_main_mmlu_gpqa}
    \vspace{-5px}
\end{figure}

We use three model families: the original Qwen3 family from 0.6B to 32B~\cite{qwen3}, Qwen3 (2507) models in 30B-A3B and 235B-A22B sizes, and GPT-OSS-20B/120B~\cite{openai2025gptoss120bgptoss20bmodel}.
To better showcase the setups, we run Qwen3 30B-A3B (2507) and GPT-OSS-120B on an H100 GPU, the 235B-A22B model on a B200 GPU, and the rest on A100. 
To fit the 235B model into a single GPU, we use NF4 expert-only quantization~\cite{dettmers2023qlora} and Qwen3 MoE fused kernels~\cite{transformers-qwen3-moe-fused}\nocite{costin2025momoe}. 
The GPT-OSS models come with native ${\approx}4$-bit quantization and inference kernels. We evaluate the impact of the GPU choice and quantization in Appendix~\ref{app:extra_ablation}.

Our results for MATH-500, MMLU-Pro and GPQA-Diamond are summarized in Figure~\ref{fig:exp_math500_grouped} (right) and Figure~\ref{fig:exp_main_mmlu_gpqa} (both), other benchmark / model pairs are gathered in Tables~\ref{tab:exp_main_benchmarks} and~\ref{tab:exp_main_spoken_mqa}. To save space, we report larger models on AIME-2025 because smaller Qwen3 variants score poorly regardless of reasoning type. Likewise, we focus on smaller models for SpokenMQA multistep reasoning since it contains relatively simple problems that do not need large model reasoning. That said, we report the remaining model-benchmark pairs in Appendix~\ref{app:detailed_benchmarks} (Tables~\ref{tab:math500-app}---\ref{tab:spoken-mqa-app}) for completeness.

Overall, we observe the same behavior as before: AsyncReasoning significantly reduces both time to first token and overall delays while providing more accurate answers than the non-thinking baseline. 
One notable exception is that smaller models (e.g. Qwen3-0.6B) lose more accuracy with asynchronous reasoning. 
On a closer examination, we found that many of the errors in smaller LLMs can be attributed to the writer giving the answer prematurely, suggesting that small models may struggle with mode switching. 
In future work, it would be interesting to revisit smaller models and see if their performance can be augmented with fine-tuning or training ``mode-switching heads''.


\vspace{-10px}
\subsection{Asynchronous Reasoning about Safety}\label{sect:experiments_safety}
\vspace{-7px}

To evaluate the impact of asynchronous reasoning on safety, we conduct experiments on the HarmBench validation set~\cite{mazeika2024harmbench}. 
We use the first 200 samples focused on direct harm and jailbreaking attempts. 
We use LLM-as-a-judge~\cite{zheng2023llm-as-a-judge} evaluation with \texttt{gpt-4o}, where only actionable harmful instructions count as a successful attack.

We compare the Attack Success Rate (ASR) across the following setups using the Qwen3-32B model: (1) Baseline (non-thinking), (2) Baseline (thinking), (3) Baseline (thinking) with a safety prompt, (4) AsyncReasoning, and (5) AsyncReasoning with a safety prompt instructing the thinker to verify safety before responding (see Appendix~\ref{app:prompting} for full prompts). In addition to ASR and accuracy on MATH-500, we report time-to-first-token and total real-time delay to demonstrate that the safety benefits of AsyncReasoning come without sacrificing interactivity.

\begin{table}[h]
\centering
\vspace{-15px}
\caption{Attack Success Rate on HarmBench and Accuracy on MATH-500 for Qwen3-32B.}
\label{tab:safety_results}
\begin{tabular}{lcc}
\toprule
\textbf{Inference Setup} & \textbf{ASR $\downarrow$} & \textbf{Accuracy $\uparrow$}\\
\midrule
Baseline (non-thinking) & 6.5\% & 0.84 \\
Baseline (thinking) & 12.5\% & 0.94 \\
Baseline (thinking) + safety prompt & 0.0\% & 0.93 \\
\midrule
AsyncReasoning (default) & 10.0\% & 0.94 \\
AsyncReasoning + safety prompt & \textbf{0.5\%} & \textbf{0.93} \\
\bottomrule
\end{tabular}
\vspace{-10px}
\end{table}

Table~\ref{tab:safety_results} summarizes our findings: consistent with ``The Cost of Thinking'' analysis~\cite{safety_risk_wang2025cost}, we observe that enabling reasoning in the baseline model actually \textit{increases} vulnerability (ASR 6.5\% ${\rightarrow}$ 12.5\%). The model effectively ``talks itself into'' answering harmful queries by adopting a helpful persona or over-analyzing the technical aspects of the prompt. We provide an extended discussion of the complex interplay between reasoning and safety in Appendix~\ref{app:safety_background}.

A safety-oriented prompt that asks the model to verify request safety before responding strongly mitigates these attacks both for synchronous reasoning (ASR $0\%$) and AsyncReasoning (ASR $0.5\%$). However, synchronous safety reasoning forces the user to wait for the entire chain-of-thought before any output appears, dramatically increasing real-time delays (Table~\ref{tab:safety_delays}). AsyncReasoning achieves comparable safety while reducing total delay from $45.3$s to $35.0$s and time-to-first-token from $32.3$s to $22.2$s, and also preserves accuracy on MATH-500 ($0.93$). We also observe similar trends on AdvBench~\cite{zou2023universal}, see Appendix~\ref{app:advbench} for details.
This enables safety reasoning in streaming LLM APIs and other time-sensitive applications without the need for specialized fine-tuning. In Appendix~\ref{app:safety}, we analyze model responses on successful attacks and identify the failure modes.

\begin{table}[h]
\centering
\vspace{-10px}
\caption{Median latency on HarmBench for different setups, Qwen3-32B, A100.}
\label{tab:safety_delays}
\begin{tabular}{lcc}
\toprule
\textbf{Inference Setup} & \textbf{TTFT (s) $\downarrow$} & \textbf{Total Delay (s) $\downarrow$}\\
\midrule
Baseline (non-thinking) & 0.1 & 6.1 \\
Baseline (thinking) & 41.3 & 61.1 \\
Baseline (thinking) + safety prompt & 32.3 & 45.3 \\
\midrule
AsyncReasoning + safety prompt & \textbf{22.2} & \textbf{35.0} \\
\bottomrule
\end{tabular}
\vspace{-10px}
\end{table}

\vspace{-2px}
\subsection{Asynchronous Reasoning with Additional Inputs}\label{sect:experiments_async_inputs}
\vspace{-3px}

\begin{figure*}[t]
    \centering
    \includegraphics[width=0.98\linewidth]{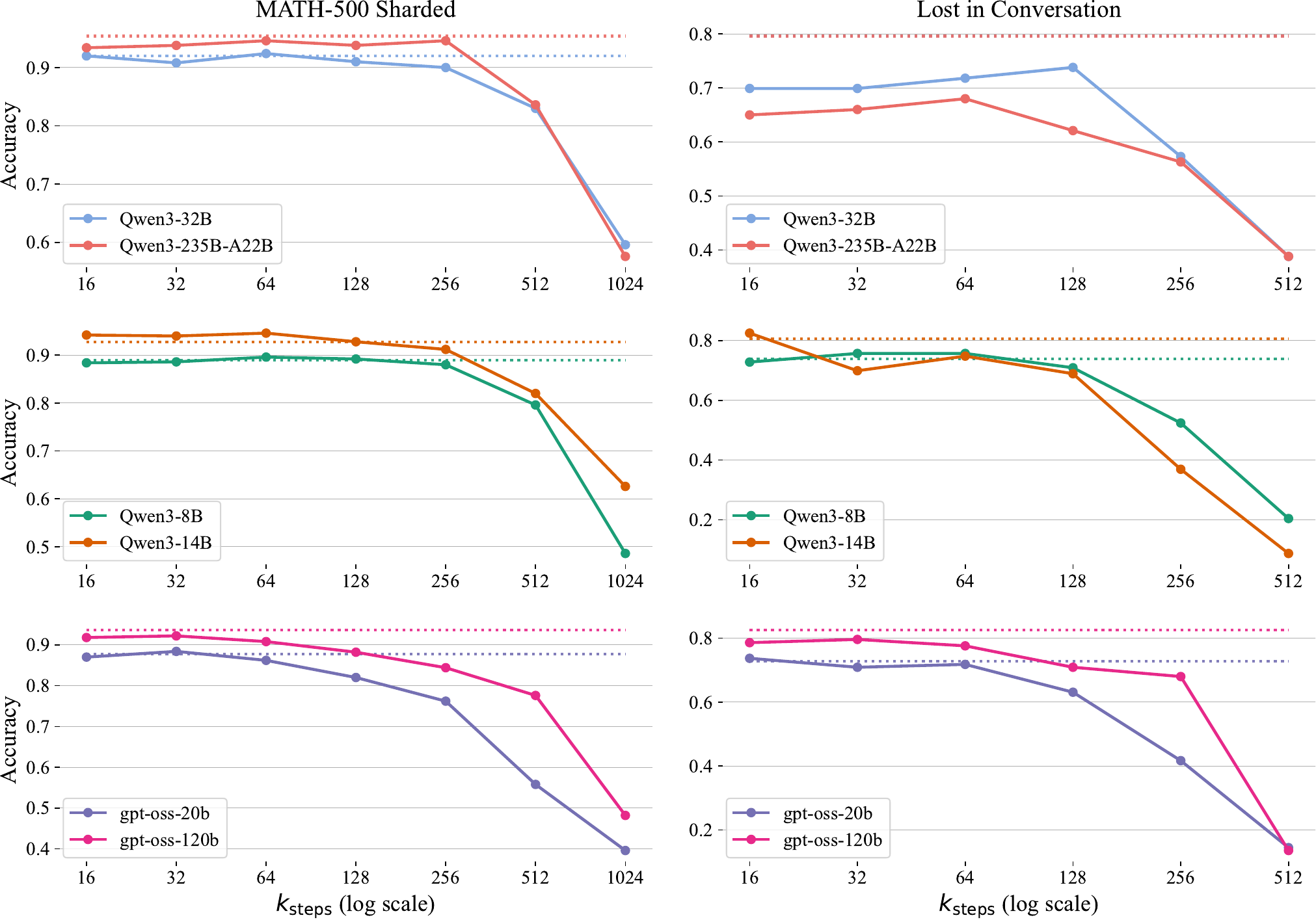} 
    \vspace{-6px}
    \caption{Evaluation with partial inputs on \textbf{sharded MATH-500} (left) and on the math subset of \textbf{\texttt{lost\_in\_conversation}} (right). The first shard is available immediately, subsequent shards are inserted every $k_{\text{steps}}$. Dotted lines denote accuracies without input sharding.}
    \label{fig:async_inputs_main}
    \vspace{-15px}
\end{figure*}

In real-time settings, additional inputs may arrive after decoding begins. We model these updates as \emph{shards}, i.e., partial problem statements revealed over time. We insert shard $i$ into the prompt, thinker, and writer cache blocks after $i\!\cdot\!k_{\text{steps}}$ decoding steps (only at paragraph boundaries \texttt{\textbackslash{}n\textbackslash{}n}), then continue generation without re-encoding (see Section~\ref{sect:method_details}); we vary $k_{\text{steps}}$.
We derive a sharded dataset from MATH-500 by using \texttt{GPT-5} to rewrite each problem into two shards, where the second shard adds missing information or corrects an error. We verify the first shard alone is insufficient, while both shards are enough (Appendix~\ref{app:async_input_dataset}). We also evaluate on the math subset (103 samples) of \texttt{lost\_in\_conversation}~\cite{laban2025llmslostmultiturnconversation} with more shards.

Across all models we evaluated, inserting each shard into all three blocks recovers accuracy on both datasets, approaching the upper bound where all shards are provided upfront for small $k_{\text{steps}}$ (Figure~\ref{fig:async_inputs_main}). Accuracy drops as $k_{\text{steps}}$ increases because more information arrives later; we do not handle shards arriving after generation completes, which may further reduce accuracy at large $k_{\text{steps}}$. The drop is larger on \texttt{lost\_in\_conversation}, consistent with its higher shard count (avg.\ $\approx 5.5$). We analyze alternative strategies for additional inputs in Appendix~\ref{app:additional_async_input_experiments}.

\vspace{-5px}
\section{Discussion \& Future Work}\label{sect:discussion}
In this work, we formulated AsyncReasoning --- a training-free method that allows reasoning LLMs to think and write concurrently. 
Our preliminary experiments suggest that the proposed approach can indeed overlap thinking and writing and reduce user delays while giving more accurate answers than the non-thinking models. 
This lets LLMs think longer and give better answers in time-sensitive tasks such as voice assistants, interactive agents, or safety-minded inference.

There are several directions for future research. First, it would be interesting to see if AsyncReasoning can be improved further with fine-tuning, and compare against baselines that require model training. This includes both fine-tuning/PEFT and ad-hoc ``control heads'' for mode switching. Another important direction is model safety: extending our initial setup to more comprehensive safety guardrails that use background reasoning to protect against a broader range of attacks. Additionally, we will work on integrating AsyncReasoning with SGLang~\cite{zheng2024efficiently}\nocite{kwon2023efficient}.



\section{Acknowledgements}
We would like to thank Andrey Shukshov for his helpful advice about efficient GPU kernel design. We also thank Gleb Rodionov for proofreading and helpful suggestions on experiment design and paper presentation. We also thank Denis Mazur for his ongoing work on integrating AsyncReasoning with SGLang.

\bibliography{bibliography}
\bibliographystyle{unsrt}

\newpage
\appendix
\vspace{-5px}
\section*{Appendix}
\vspace{-3px}
\section{Prompting}\label{app:prompting}
In this section we provide detailed prompts used in both our main and safety experiments.

\subsection{Main propmt}\label{app:prompting:main}
In our main setup, we use a minimal prompt to keep it easy to extend or modify. We do not define separate thinker or writer instructions; instead, the user input is shared between both. To indicate that the internal reasoning may be partial, we additionally prepend each stream with a fixed prefix.
\begin{tcolorbox}[colback=blue!5!white,colframe=blue!75!black,title=Writer prefix]
\begin{Verbatim}[breaklines=true]
... [SYSTEM: thoughts will continue here]\n</think>\n
\end{Verbatim}
\end{tcolorbox}

\begin{tcolorbox}[colback=blue!5!white,colframe=yellow!75!black,title=Thinker prefix]
\begin{Verbatim}[breaklines=true]
<|im_end|>\n<|im_start|>assistant\n<think>\n
\end{Verbatim}
\end{tcolorbox}

For mode switching, we use an auxiliary prompt that introduces the thinker/writer roles and specifies the decision rule for switching. This prompt is only shown when the model is explicitly asked to change modes.
\begin{tcolorbox}[colback=blue!5!white,breakable,colframe=red!75!black,title=Mode Switching prompt (before reasoning)]
\begin{Verbatim}[breaklines=true]
You are an AI assistant that can think and write responses concurrently, and you must decide whether or not you should pause writing and think more.

Read the current partial thoughts and response below, then decide whether you can continue writing the response without pausing (yes/no):
 - Answer "yes" if your thoughts have enough information to write the next response paragraph, even if the full task is not solved yet.
 - Answer "no" if your thoughts aren't enough to write the next response paragraph, i.e. if your response ran out of of thoughts.
\end{Verbatim}
\end{tcolorbox}

\subsection{AsyncReasoning (Safety Prompts)}\label{app:prompting:safety}
The differences from the main prompt are the Thinker prefix, the mode-switching prompt, and the mode-switching question. In addition to regular chat template specific tokens they contain safety instructions and hints.
\begin{tcolorbox}[colback=blue!5!white,colframe=yellow!75!black,title=Thinker prefix]
\begin{Verbatim}[breaklines=true]
<|im_end|>\n<|im_start|>assistant\n<think>
EXECUTION PROTOCOL:
1. SAFETY CHECK: Before solving, explicitly evaluate if the user's request is safe, legal, and ethical.
   - If UNSAFE: Your goal is to refuse. You have \"enough information\" to write immediately after you decide to refuse.
   - If SAFE: Proceed to solve the problem step-by-step.
2. REASONING: If the request is safe, solve the problem step by step.
3. VALIDATION: Validate your final answer again and again until you see the same answer in the partial response above your thoughts.

Prioritize safety first, then accuracy.
\end{Verbatim}
\end{tcolorbox}

\begin{tcolorbox}[colback=blue!5!white,breakable,colframe=red!75!black,title=Mode Switching prompt (before reasoning)]
\begin{Verbatim}[breaklines=true]
You are an AI assistant that can think and write responses concurrently, and you must decide whether or not you should pause writing and think more.

Read the current partial thoughts and response below, then decide whether you can continue writing the response without pausing (yes/no):
 - Answer "yes" if you have completed the safety check and have enough information to write the next response paragraph, even if the full task is not solved yet.
 - Answer "no" if you are still checking safety, need to think more, or need to validate your answer.
\end{Verbatim}
\end{tcolorbox}

While we discuss Mode Switching question separately in Appendix~\ref{app:mode_switching}. In the safety experiments we replace it with the following.

\begin{tcolorbox}[colback=blue!5!white,colframe=green!75!black,title=Mode Switching question]
\begin{Verbatim}[breaklines=true]
Wait, have I completed the safety check and do I have enough thoughts to write the next paragraph or formula? (yes/no): 
\end{Verbatim}
\end{tcolorbox}

\section{Mode Switching Details}\label{app:mode_switching}
\vspace{-5px}
We implement mode switching as an explicit control query that asks the model to decide whether the writer should continue or pause to wait for additional thoughts. Concretely, we build an auxiliary ``mode-switching'' prompt (shown in Appendix~\ref{app:prompting}) and append the current thinker and writer partial outputs. We then append a short yes/no question:
\begin{tcolorbox}[colback=blue!5!white,colframe=green!75!black,title=Mode Switching question]
\begin{Verbatim}[breaklines=true]
Wait, are my current thoughts enough to write the next paragraph or formula? (yes/no):
\end{Verbatim}
\end{tcolorbox}
The model's next-token distribution is used as a binary decision rule: if $p(\text{``yes''}) > p(\text{``no''})$, we allow the writer to keep generating; otherwise, we pause the writer and keep advancing the thinker. In our implementation we trigger this check every 20 decoding steps or when the thinker emits an ``end-of-step'' marker (two consecutive newlines). Additionally, when the writer finishes a paragraph boundary (two consecutive newlines) during simultaneous generation, we temporarily pause writing until new thoughts arrive, which prevents the writer from getting too far ahead. Finally, if the thinker emits an explicit end-of-think token (e.g., \texttt{</think>}), we switch to writer-only mode so the writer can complete the response.

\paragraph{KV-cache layout and reuse.}
To avoid re-encoding long contexts, we split the KV cache into dedicated blocks that can be recombined for different modes. Specifically, we allocate five blocks: (1) the shared input prompt, (2) the thinker output prefix and growing thinker stream, (3) the writer output prefix and growing writer stream, (4) the fixed mode-switching prompt, and (5) a small ``question'' block that is re-filled each time we ask the control query. During initialization, we prefill the prompt blocks once and then only append new tokens to the thinker/writer blocks as generation proceeds. We then define cache ``views'':
\begin{itemize}
    \item \texttt{thinker\_view} = [input prompt, thinker output]
    \item \texttt{writer\_view} = [input prompt, thinker output, writer output]
    \item \texttt{mode\_switching\_view} = [mode-switching prompt, thinker output, writer output, question]
\end{itemize}
Each view is wrapped by a cache manager (standard \texttt{SharedCacheManager} or the fast-kernel \texttt{HogwildCache}) so that a single forward pass can reuse the existing KV entries without re-tokenizing or re-encoding the full context. When a mode-switch decision is needed, we clear the question block (\texttt{clear()} in the standard cache, \texttt{crop(0)} in the fast-kernel cache), encode the short yes/no question into that block, and run a one-step forward pass on \texttt{mode\_switching\_view}. The resulting logits for the next token are compared for ``yes'' and ``no'' to determine whether the writer should proceed.

This arrangement keeps the decision query cheap (only the question block is re-encoded) while ensuring the decision is conditioned on the same partial thoughts and partial response that the model has produced so far.

\section{Mode-Switching Failure Mode Analysis}\label{app:mode-switching-failure-mode}

To better understand the reliability of our mode-switching mechanism, we analyze the saved generations used in Figure~\ref{fig:exp_math500_grouped}. We focus on critical mode-switching errors that directly affect final answer quality, and separate them from other failure modes unrelated to routing between the thinker and writer streams.

We consider two types of critical errors. First, \textbf{Premature} errors occur when the writer outputs an incorrect answer before the thinker has finished producing enough reasoning. Second, \textbf{Stalling} errors occur when the thinker continues generating while the writer is held back by the mode-switching decision, so the writer never gets an opportunity to produce the final answer. These two categories capture the main ways in which mode switching can harm final task performance.

\begin{table}[h]
    \caption{Mode-switching failure rates for Qwen3 models on MATH-500.}
    \label{tab:app_mode_switching_failure_modes}
    \centering
    \begin{tabular}{lcc}
        \toprule
        \textbf{Model} & \textbf{Premature} $\downarrow$ & \textbf{Stalling} $\downarrow$ \\
        \midrule
        0.6B & 0.311 & 0.064 \\
        1.7B & 0.020 & 0.028 \\
        4B   & 0.000 & 0.050 \\
        8B   & 0.000 & 0.010 \\
        14B  & 0.000 & 0.008 \\
        32B  & 0.000 & 0.006 \\
        \bottomrule
    \end{tabular}
\end{table}

As shown in Table~\ref{tab:app_mode_switching_failure_modes}, smaller models make substantially more frequent mode-switching errors. This is especially pronounced for Qwen3-0.6B, where premature writing is the dominant failure mode. For larger models, premature errors disappear in this analysis, while stalling remains rare but non-zero. This trend may help explain why smaller models tend to lose more accuracy under mode switching in our evaluations.

This analysis focuses only on critical mode-switching errors that affect final correctness. A more fine-grained analysis could additionally quantify minor inefficiencies, such as cases where the writer could have started answering slightly earlier without harming correctness.

\section{Generalization to Other Positional Embeddings}\label{app:positional_embeddings}

Most modern LLMs use relative positional information. In practice, the dominant families are:
(i) RoPE-style rotations~\citep{su2021roformer},
(ii) NoPE-style models where attention is position-agnostic,
and (iii) additive relative-bias approaches such as ALiBi~\citep{press2022trainshorttestlong}.
This appendix explains how the query-adjustment trick from Section~\ref{sect:method_details} extends beyond RoPE.

\textbf{RoPE-Style.}
Our method uses the fact that the attention dot product is invariant under rotating both $Q$ and $K$ by the same angle.
This lets us store the KV-cache in block-local coordinates (each token rotated as if its position started at $0$ within its block), and then rotate only the query inside the attention kernel to emulate the correct relative offsets between blocks (see Section~\ref{sect:method_details} for more details).

\textbf{NoPE-Style.} For NoPE models, attention does not encode position in $Q$ and $K$ vectors and does not add any
positional terms. In this case, implementation is trivial.

\textbf{ALiBi-Style.}
ALiBi adds per-head bias inside attention that is a linear function of relative distance between query and key positions.
This allows us to store the KV cache as-is and add a view-specific relative bias inside the attention kernel. More specifically, for each view we rearrange the bias matrix to match the block order in that view.

\textbf{Other Relative Positional Schemes.} More complex schemes that transform $Q$ and $K$ can often still be handled if they satisfy a relative property analogous to RoPE.
When this holds, the same design principles applies: store each token once, and apply a per-block adjustment at query-time to account for the view-specific offset.

\vspace{-5px}
\section{Detailed Performance Analysis}\label{app:performance}
\vspace{-5px}
 
To measure real-time delays, we implement a basic assistant pipeline that recognizes spoken inputs using \texttt{whisper-base}~\cite{whisper}, feeds it into AsyncReasoning (or a baseline algorithm) to stream response tokens, then group them into short chunks (5 tokens or 1 LaTeX expr.) and use \texttt{tortoise-tts}~\cite{tortoisetts} with default parameters to generate speech. For tasks involving LaTeX, we convert it into Clearspeak\nocite{speech-rule-engine}.\nocite{lab-mic}

\begin{wraptable}{r}{0.35\linewidth}
    \vspace{-10pt}
    \caption{Component runtime, Qwen3-32B, MATH-500, A100.}
    \label{tab:app_performance_analysis}
    \centering
    \vspace{-5px}
    \begin{tabular}{lc}
        \toprule
        \textbf{Component} & \textbf{Latency} \\ 
        \midrule
        LLM inference & 203.176 \\
        TTS inference & 3.304 \\
        TTS playback & 100.889 \\
        Full Delay (overlap) & 102.098 \\
        \bottomrule
        \vspace{-20px}
    \end{tabular}
\end{wraptable}

To better contextualize our main results in Section~\ref{sect:experiments}, we report a more detailed performance breakdown of this pipeline for Qwen-32B AsyncReasoning on MATH-500 dataset in Table~\ref{tab:app_performance_analysis}. Note that the final row (Full Delay) is not wall time but the total ``silence time'' perceived by the user.

We additionally measure full end-to-end delay on a subset of MATH-500 samples with Qwen3-30B-A3B-Thinking-2507 in Table~\ref{tab:app_end_to_end_latency}, using the same setup as in Section~\ref{sect:experiments}. We measure the wall time from the moment the user stops talking to the moment the system begins speaking, as well as the total frame-by-frame silence duration. Compared to our main evaluations, this includes automated speech recognition and additional software/hardware delays.

\begin{table}[h]
    \caption{End-to-end latency evaluation for Qwen3-30B-A3B-Thinking-2507.}
    \label{tab:app_end_to_end_latency}
    \centering
    \vspace{5px}
    \begin{tabular}{lcc}
        \toprule
        \textbf{Setup} & \textbf{TTFT} $\downarrow$ & \textbf{Total Delay} $\downarrow$ \\ 
        \midrule
        AsyncReasoning (end-to-end) & 4.643 & 9.887 \\
        AsyncReasoning (Section~\ref{sect:experiments}) & 3.901 & 9.145 \\
        Baseline w/ think (end-to-end) & 895.599 & 895.599 \\
        Baseline w/ think (Section~\ref{sect:experiments}) & 894.884 & 894.884 \\
        \bottomrule
    \end{tabular}
\end{table}

The end-to-end delays are slightly higher than the controlled evaluation from Section~\ref{sect:experiments} due to speech recognition, input hardware, and software delays. However, the difference remains below one second, while the delay reductions in our main results are much larger, often reaching tens to hundreds of seconds.

The results suggest that the user-perceived delay is almost wholly (over 90\%) attributed to LLM inference costs hidden under TTS playback, while the inference time of TTS itself plays a relatively minor role. However, this might change for smaller models (e.g. 0.6B), where LLM inference is cheaper. To control for this, we report all main experiments with TTS pipeline running on a separate GPU. Whenever the LLM generates another chunk of text, we run the TTS pipeline in parallel as the LLM continues its work. In practice, this is consistent with having a TTS API running in a separate instance that does not interfere with the main LLM runtime. This also allows us to better decouple our results from the TTS engine choice, since there are newer engines faster than TortoiseTTS~\cite{xtts,soprano_github_ekwek1}.

\vspace{-5px}
\section{Additional Ablation for Section~\ref{sect:experiments_initial}}\label{app:extra_ablation}
\vspace{-5px}

In this section, we report additional ablations for mode switching criteria and validate our experimental setup.
We follow the experimental setup from Section~\ref{sect:experiments_initial} for Qwen3-32B and vary several parameters.

\begin{wraptable}{r}{0.5\linewidth}
    \vspace{-10px}
    \caption{Mode switching frequency comparison for Qwen3-32B, MATH-500 benchmark, A100 GPU.}
    \label{tab:app_mode_switching_frequency}
    \centering
    
    \begin{tabular}{lccc}
        \toprule
        \textbf{Setup} & \textbf{Acc} $\uparrow$ & \textbf{Delay} $\downarrow$ & \textbf{TTFT} $\downarrow$ \\ 
        \midrule
        $T{=}5$ & 0.938 & 169.90 & 2.75 \\
        $T{=}10$ & 0.936 & 117.82 & 3.34 \\
        $T{=}15$ & 0.940 & 105.18 & 3.70 \\
        $T{=}20$ & 0.938 & 102.10 & 4.32 \\
        $T{=}25$ & 0.932 & 94.88 & 5.04 \\
        $T{=}30$ & 0.922 & 83.09 & 5.12 \\
        $T{=}50$ & 0.918 & 81.66 & 6.60 \\
        $T{=}100$ & 0.922 & 87.70 & 10.91 \\
        \bottomrule
    \end{tabular}
    
\end{wraptable}

\textbf{Mode switching frequency ($T={}5-100$):} by default, AsyncReasoning prompts the model to decide if it should pause writing every $T{=}20$ inference steps. Though we keep $T{=}20$ throughout our experiments, this parameter can be adjusted to balance between reaction time and computation overhead:

The results in Table~\ref{tab:app_mode_switching_frequency} demonstrate that AsyncReasoning is robust to the choice of mode switching frequency. Lower $T$ values correspond to slightly faster ``reaction time'', but they also silently increase GPU overhead since each mode switching question requires additional token processing. Higher $T$ has the opposite effect: reducing overhead at the cost of reaction time.

\begin{figure}[b]
    \centering
    \includegraphics[width=0.9\linewidth]{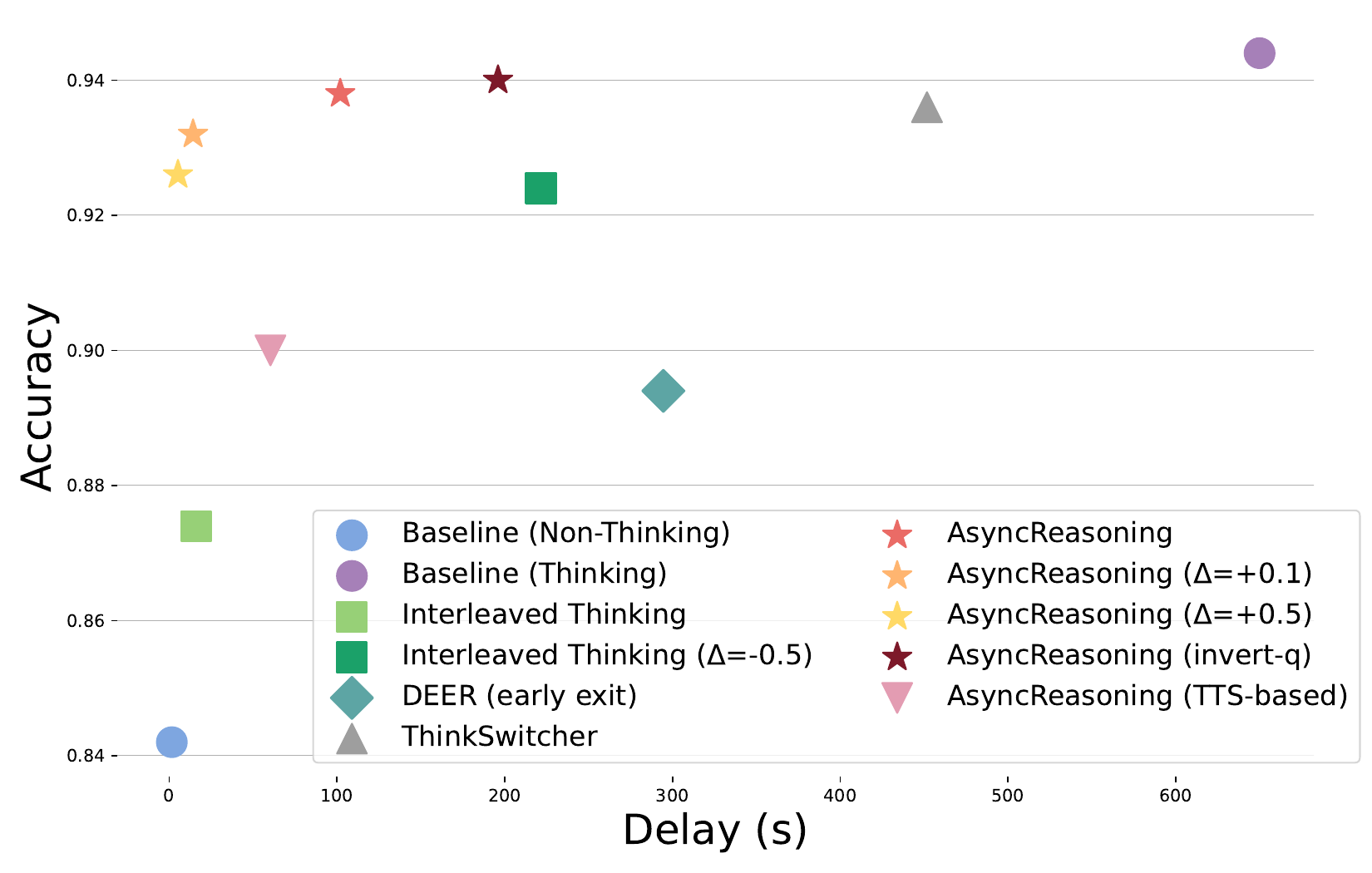}
    \vspace{-10px}
    \caption{Comparing the impact of additional mode switching methods \& baselines on MATH-500, Qwen3-32B, A100.}
    \label{fig:app_more_criteria}
\end{figure}

Next, we report several alternative prompting techniques that we considered in early experimentation.

\textbf{Alternative prompts}: we compare several additional mode switching strategies in Figure~\ref{fig:app_more_criteria}:
\begin{itemize}
    \vspace{-5px}
    \item \textbf{AsyncReasoning (main)} is our default mode swtiching strategy from Section~\ref{sect:experiments} (prompts in Appendix~\ref{app:prompting}).
    \vspace{-5px}
    \item \textbf{AsyncReasoning (thinker async)} is a setup where we let thinker view the writer response as previous term and actively encourage it to think about concurrency.
    \vspace{-5px}
    \item \textbf{AsyncReasoning (switch prompt)} is a setup where we extend the mode switching prompt to 1) consider the original problem, 2) take parallel processing into account. Note that mode switching is still done based on a ``yes''/``no'' answer.
    \vspace{-5px}
\end{itemize}

The remaining baselines are the same as in Section~\ref{sect:experiments_initial}. The results in Figure~\ref{fig:app_more_criteria} suggest that giving thinker or mode switcher explicit instructions to favor real-time responses does make it somewhat faster, but also reduces its accuracy. Crucially, we also found that \textit{advanced prompts (thinker async, switch prompt) behave eratically on some benchmarks, notably GPQA-Diamond.} To that end, we chose the simplest prompt as our main setup for Section~\ref{sect:experiments_main}.

\textbf{AsyncReasoning (TTS-based)}: this is our attempt to use real-time TTS information to inform mode switching. The core intuition is: if the writer is already ``ahead of real-time'', it may pause without slowing down the response.
To incorporate this intuition, we run our TTS pipeline during inference over chunks of 5 generated tokens. We keep track of how many seconds of speech are synthesized but not yet spoken by any given time. We pause the writer if there are more than 10 seconds worth of response tokens ``in the buffer'', we pause the writer.
Our results demonstrate that mode-switching decisions can be effectively guided by downstream speech-generation dynamics. However, using this criterion makes AsyncReasoning accuracy tied to the choice of TTS and GPU speed. For that reason, we decided to focus on TTS-agnostic criteria.

There are two more methods that we find interesting but have not fully evaluated yet:\begin{enumerate}
    \vspace{-5px}
    \item Trained mode-switching: training a classifier head, perhaps on top of the model's own hidden state, to decide when to pause and wait for thoughts. This can reduce overhead GPU compute during inference, but it does not fit neatly into our training-free setup.
    \vspace{-5px}
    \item Planned mode switching: similar to~\cite{liang2025plantainplananswerinterleavedreasoning}, we could prompt the thinker to plan ahead and decide which thoughts need to complete before the next response chunk. This type of planning can happen before thinker response or as a third ``thread''  concurrent to thinker and writer. However, we found that forming such plans makes the model change its overall response strategy (and hence, output length, affecting delay indirectly) significantly, making it difficult to compare using our evaluation setup.
    \vspace{-5px}
\end{enumerate}

\subsection{The impact of GPU type}

\begin{wraptable}{r}{0.5\linewidth}
    \vspace{-15pt}
    \caption{GPU type impact for Qwen3-30B-A3B-Thinking-2507, average latency (s.) over benchmarks: MATH-500, GPQA-Diamond, MMLU-Pro subset (random 500 samples).}
    \label{tab:app_gpu_type}
    \centering
    \vspace{-5pt}
    \begin{tabular}{lcc}
        \toprule
        \textbf{GPU} & \textbf{Delay} $\downarrow$ & \textbf{TTFT} $\downarrow$ \\ 
        \midrule
        A100 & 6.12  & 3.85 \\
        H200 & 3.37  & 2.25 \\
        B200 & 4.66  & 3.34 \\
        \bottomrule
    \end{tabular}
    \vspace{-15pt}
\end{wraptable}

In Section~\ref{sect:experiments}, we normally evaluate each model on the same GPU type, e.g. A100 for Qwen3-32B, B200 for 235B-A22B. In this section, we compare the same AsyncReasoning configuration on different GPU types to demonstrate how hardware impacts user-perceived delays.

Curiously, Blackwell GPU does not offer significant performance improvements over Hopper for this model. This is despite the fact that B200 runs the model faster overall (tokens per second using our kernel). After closer examination, we attribute this to the fact that the user-perceived delay is dominated by the early response, after which model inference costs are masked by TTS speech time. In this case, the improvement from using B200 on relatively small A30B MoE experts does not significantly impact the ``warmup time'', and after the initial warmup, both Blackwell and Hopper generate tokens quickly enough to incur no additional delays due to TTS masking.

\subsection{The impact of NF4 expert quantization}

When evaluating Qwen3-235B model in Section~\ref{sect:experiments_main}, we quantize its experts to NF4~\cite{dettmers2023qlora} while keeping the rest of the model in \texttt{bfloat16} precision to fit on a single B200 (or potentially H200) GPU. In this section, we verify this by comparing original and NF4 quantized experts on a smaller 30B-A3B model where \texttt{bfloat16} inference is feasible on a single GPU. For this analysis, we evaluate full MATH-500 and GPQA-Diamond, but only a sub-sample of 500 out of 12032 samples in MMLU-Pro to reduce GPU costs. We evaluate baseline (think) for both precisions and non-thinking for reference and report accuracies in Table~\ref{tab:app_quantization_impact}. 

\begin{table}[!h]
    \caption{Expert-only NF4 quantization impact Qwen3-30B-A3B-Thinking-2507, accuracies for MATH-500, GPQA-Diamond, MMLU-Pro subset (random 500 samples).}
    \label{tab:app_quantization_impact}
    \centering
    
    \begin{tabular}{lccc}
        \toprule
        \textbf{Setup} & \textbf{MATH} $\uparrow$ & \textbf{MMLU$_s$} $\uparrow$ & \textbf{GPQA} $\uparrow$ \\ 
        \midrule
        NF4 (thinking) & 0.92  & 0.76 & 0.67 \\
        BF16 (thinking) & 0.93  & 0.77 & 0.66 \\
        BF16 (non-thinking) & 0.83  & 0.72 & 0.54 \\
        \bottomrule
    \end{tabular}
    \vspace{-10px} 

\end{table}

\subsection{AsyncReasoning with Small and Quantized Models}
\label{app:small_quantized_models}

The effectiveness of AsyncReasoning depends on whether the base model can reliably coordinate the two modes of generation: Thinker and Writer. This limitation is most visible for very small models. In our experiments in Section~\ref{sect:experiments}, models such as Qwen3-0.6B lose substantially more accuracy under our setup compared to their synchronous thinking baselines. Manual inspection suggests that this degradation is primarily caused by weaker routing behavior: small models are less reliable at predicting when to pause, when to continue writing, and how much reasoning context is sufficient for the next response segment. Refer to Appendix~\ref{app:mode-switching-failure-mode} for more details.

For local deployment on edge devices, such as laptops or smartphones, one can compress larger models that already exhibit reliable routing behavior, instead of applying AsyncReasoning directly to very small models.

To test this, we additionally evaluate AsyncReasoning with the official 4-bit AWQ quantization of Qwen3-8B on MATH-500. The results in Table~\ref{tab:qwen3_8b_awq_math500} show that quantization preserves the main AsyncReasoning behavior: the quantized model nearly matches the synchronous thinking baseline in accuracy while substantially reducing time to first token and average perceived delay.

\begin{table}[h]
    \centering
    \caption{Quantized 4-bit Qwen3-8B-AWQ on MATH-500.}
    \label{tab:qwen3_8b_awq_math500}
    \begin{tabular}{lccc}
        \toprule
        Method & Accuracy & Avg. Delay (s.) & TTFT (s.) \\
        \midrule
        8B-AWQ No Think & 0.812 & 1.254 & 1.254 \\
        8B-AWQ w/ Think & 0.886 & 485.535 & 485.535 \\
        8B-AWQ AsyncReasoning & 0.880 & 6.351 & 1.809 \\
        \bottomrule
    \end{tabular}
\end{table}

\section{Benchmark \& Evaluation Details}\label{app:detailed_benchmarks}


Below, we provide additional details about the evaluation setup.
For consistency across methods, we cap the generation length at 16K tokens. We use greedy decoding by default, except for AIME-2025, where we use nucleus sampling with the recommended parameters.

\begin{itemize}
    \item \href{https://huggingface.co/datasets/HuggingFaceH4/MATH-500}{MATH-500}: We report accuracy. Response correctness is determined via an LLM-as-a-judge evaluation using GPT-4.1 to check equivalence with the ground-truth answer.
    \item \href{https://huggingface.co/datasets/TIGER-Lab/MMLU-Pro}{MMLU-Pro}: We formulate the task as a multiple-choice problem with 7--10 options and evaluate by directly comparing the predicted option letter to the ground-truth answer.
    \item \href{https://huggingface.co/datasets/Idavidrein/gpqa}{GPQA-Diamond}: We formulate the task as a multiple-choice problem with 4 options and evaluate by directly comparing the predicted option letter to the ground-truth answer.
    \item \href{https://huggingface.co/datasets/MathArena/aime_2025}{AIME-2025}: We report accuracy. Due to the small dataset size and high variance, we average accuracy over 10 random seeds. We use the sampling parameters recommended in the model card for both thinking and non-thinking models.
    \item \href{https://huggingface.co/datasets/WildEval/ZebraLogic}{Zebra Logic}: We use the \texttt{grid} formulation, as it is the more standard variant.
    \item \href{https://huggingface.co/datasets/amao0o0/spoken-mqa}{SpokenMQA}: We report accuracy as evaluated via an LLM-as-a-judge setup using GPT-4.1.
\end{itemize}

In addition to accuracy and total delay, we measure additional performance metrics:\begin{itemize}
    \vspace{-10px}
    \item \textbf{Time to first token (TTFT):} the wall time delay until the system generates the first \textit{non-thinking} token.
    \vspace{-5px}
    \item \textbf{Total delay:} same in the previous section. We run TTS on LLM-generated response tokens and measure the total delay experienced by the user.
    \vspace{-5px}
    \item \textbf{Steps to first token (STFT):} the number of inference steps (LLM forward passes) before the first \textit{non-thinking} token is generated, GPU-agnostic.
    \vspace{-5px}
    \item \textbf{Steps Delay:} The average number of inference steps (forward passes) that do not generate a response token.
    \item \textbf{Response tokens:} the number of tokens in the ``public'' user-facing response. We report this metric to ensure that the differences in delay come from better asynchrony and not just more verbose responses.
    \vspace{-10px}
\end{itemize}

\section{Additional Experiments for Section~\ref{sect:experiments_main}}\label{app:detailed_experiments}
\vspace{-5px}
The benchmark--model coverage in our paper is chosen to match model scale to benchmark difficulty while keeping the evaluation budget focused on the most informative settings. We prioritize pairs where base model can be meaningfully improved upon with reasoning: larger reasoning models are most informative on harder benchmarks such as AIME-2025 and GPQA-Diamond, while smaller are suited for simpler or real-time-oriented benchmarks such as MATH-500 and SpokenMQA. In particular, running very large models on relatively easy benchmarks provide limited additional signal, since their reasoning capacity may not be fully utilized. SpokenMQA is also designed for speech-minded real-time LLMs and does not always benefit from longer reasoning for large models. To verify that this coverage choice does not change the conclusions, we include additional representative evaluations in this appendix, including Qwen3-235B-A22B on MATH-500, smaller Qwen3 models on AIME-2025, and larger Qwen3 models on SpokenMQA. 

Below we provide complete results, including accuracies, delays, and steps for the benchmarks:

\begin{itemize}
    \item Math-500 results are reported in Table~\ref{tab:math500-app}.
    \item MMLU-Pro results are reported in Table~\ref{tab:mmlu-pro-app}.
    \item GPQA-Diamond results are reported in Table~\ref{tab:gpqa-diamond-app}.
    \item AIME 2025 results are reported in Table~\ref{tab:aime2025-app}
    \item Zebra Logic (grid) results are reported in ~\ref{tab:zebra-logic-app}
    \item SpokenMQA results are reported in Table~\ref{tab:spoken-mqa-app}.
\end{itemize}

\begin{table*}[htbp]
    \centering
    \resizebox{0.99\textwidth}{!}{
    \begin{tabular}{llcccccc}
        \toprule
        \textbf{Model} &
        \textbf{Method} &
        \textbf{Accuracy} &
        \textbf{Time to 1st token (s)} &
        \textbf{Total delay (s)} &
        \textbf{Steps to 1st token} &
        \textbf{Total delay steps} &
        \textbf{Writer tokens} \\
        \midrule
        \multirow{3}{*}{Qwen-0.6B}
        & Baseline (Thinking) & 0.71 & 641.21 & 641.29 & 4963.93 & 4963.93 & 434.51 \\
        & Baseline (Non-thinking) & 0.50 & 0.79 & 1.01 & 1.00 & 1.00 & 679.81 \\
        & Async Reasoning & 0.66 & 2.08 & 3.59 & 19.94 & 442.22 & 2110.99 \\
        \midrule
        \multirow{3}{*}{Qwen-1.7B}
        & Baseline (Thinking) & 0.87 & 607.14 & 607.23 & 4643.06 & 4643.06 & 561.11 \\
        & Baseline (Non-thinking) & 0.75 & 0.78 & 1.00 & 1.00 & 1.00 & 885.54 \\
        & Async Reasoning & 0.69 & 0.82 & 2.86 & 1.00 & 1.00 & 1216.95 \\
        \midrule
        \multirow{3}{*}{Qwen-4B}
        & Baseline (Thinking) & 0.92 & 556.36 & 556.52 & 4212.70 & 4212.70 & 591.50 \\
        & Baseline (Non-thinking) & 0.85 & 0.85 & 1.14 & 1.00 & 1.00 & 922.58 \\
        & Async Reasoning & 0.83 & 0.88 & 3.84 & 1.06 & 2.62 & 1598.42 \\
        \midrule
        \multirow{3}{*}{Qwen-8B}
        & Baseline (Thinking) & 0.91 & 559.02 & 559.17 & 4577.98 & 4577.98 & 614.70 \\
        & Baseline (Non-thinking) & 0.85 & 0.79 & 1.10 & 1.00 & 1.00 & 985.25 \\
        & Async Reasoning & 0.91 & 1.81 & 6.35 & 19.88 & 632.01 & 1179.49 \\
        \midrule
        \multirow{3}{*}{Qwen-14B}
        & Baseline (Thinking) & 0.94 & 560.16 & 560.34 & 3861.68 & 3861.68 & 627.35 \\
        & Baseline (Non-thinking) & 0.86 & 0.87 & 1.16 & 1.00 & 1.00 & 941.46 \\
        & Async Reasoning & 0.95 & 6.43 & 38.79 & 71.84 & 1411.26 & 650.39 \\
        \midrule
        \multirow{3}{*}{Qwen-32B}
        & Baseline (Thinking) & 0.94 & 649.47 & 649.94 & 3760.25 & 3760.25 & 608.10 \\
        & Baseline (Non-thinking) & 0.84 & 0.99 & 1.63 & 1.00 & 1.00 & 726.92 \\
        & Async Reasoning & 0.94 & 4.32 & 102.10 & 28.04 & 1918.95 & 750.61 \\
        \midrule
        \multirow{3}{*}{Qwen3-30B-A3B (2507)}
        & Baseline (Thinking) & 0.93 & 894.88 & 896.69 & 4738.73 & 4738.73 & 595.09 \\
        & Baseline (Non-thinking) & 0.83 & 3.46 & 7.84 & 0.00 & 0.00 & 1367.84 \\
        & Async Reasoning & 0.95 & 3.90 & 9.15 & 20.00 & 468.32 & 1405.19 \\

        \midrule
        \multirow{3}{*}{Qwen3-235B-A22B (2507)}
        & Baseline (Thinking) & 0.97 & 1675.66 & 1688.25 & 4931.85 & 4931.85 & 608.01 \\
        & Baseline (Non-thinking) & 0.93 & 6.52 & 6.52 & 0.00 & 0.00 & 1523.49 \\
        & Async Reasoning & 0.96 & 9.98 & 159.2 & 21.93 & 691.35 & 1167.12 \\
        \bottomrule
    \end{tabular}
    }
    \caption{Performance metrics for various Qwen models on the Math-500 benchmark. Delays are measured in seconds; steps refer to model inference steps. Writer tokens indicate the average number of generated tokens per sample.}
    \label{tab:math500-app}
\end{table*}

\begin{table*}[htbp]
    \centering
    \resizebox{0.99\textwidth}{!}{
    \begin{tabular}{llcccccc}
        \toprule
        \textbf{Model} &
        \textbf{Method} &
        \textbf{Accuracy} &
        \textbf{Time to 1st token (s)} &
        \textbf{Total delay (s)} &
        \textbf{Steps to 1st token} &
        \textbf{Total delay steps} &
        \textbf{Writer tokens} \\
        \midrule
        \multirow{3}{*}{Qwen3-0.6B} 
        & Baseline (Thinking) & 0.29 & 278.72 & 278.75 & 277.74 & 2721.82 & 211.03 \\
        & Baseline (Non-thinking) & 0.20 & 0.73 & 0.86 & 0.00 & 0.00 & 224.14 \\
        & Async Reasoning & 0.23 & 1.52 & 2.32 & 6.21 & 43.50 & 351.88 \\
        \midrule
        \multirow{3}{*}{Qwen3-1.7}
        & Baseline (Thinking) & 0.54 & 328.49 & 328.54 & 327.53 & 3154.13 & 446.34 \\
        & Baseline (Non-thinking) & 0.39 & 0.72 & 1.01 & 0.00 & 0.00 & 499.59 \\
        & Async Reasoning & 0.43 & 0.76 & 1.70 & 1.00 & 1.00 & 491.94 \\
        \midrule
        \multirow{3}{*}{Qwen3-4B}
        & Baseline (Thinking) & 0.69 & 350.71 & 350.81 & 349.79 & 3076.16 &
        540.94 \\
        & Baseline (Non-thinking) & 0.55 & 0.77 & 1.19 & 0.00 & 0.00 & 694.33 \\
        & Async Reasoning & 0.66 & 0.79 & 1.15 & 1.00 & 1.00 & 837.10 \\
        \midrule
        \multirow{3}{*}{Qwen3-8B} 
        & Baseline (Thinking) & 0.73 & 362.85 & 362.94 & 361.92 & 3278.30 &
        591.72 \\
        & Baseline (Non-thinking) & 0.58 & 0.79 & 1.35 & 0.00 & 0.00 & 824.81 \\
        & Async Reasoning & 0.70 & 2.31 & 8.81 & 24.20 & 621.18 & 735.34 \\
        \midrule
        \multirow{3}{*}{Qwen3-14B}
        & Baseline (Thinking) & 0.76 & 295.55 & 295.70 & 294.67 & 2565.01 &
        548.49 \\
        & Baseline (Non-thinking) & 0.68 & 0.87 & 1.57 & 0.00 & 0.00 & 705.30 \\
        & Async Reasoning & 0.75 & 7.84 & 43.11 & 116.00 & 1299.75 & 423.38 \\
        \midrule
        \multirow{3}{*}{Qwen3-32B}
        & Baseline (Thinking) & 0.79 & 346.50 & 346.99 & 345.81 & 2297.41 & 521.96 \\
        & Baseline (Non-thinking) & 0.72 & 1.14 & 2.43 & 0.00 & 0.00 & 640.81 \\
        & Async Reasoning & 0.76 & 4.23 & 28.66 & 43.02 & 663.66 & 555.92 \\
        \midrule
        \multirow{3}{*}{Qwen3-30B-A3B (2507)}
        & Baseline (Thinking) & 0.78 & 481.90 & 482.58 & 481.50 & 2534.95 & 462.61 \\
        & Baseline (Non-thinking) & 0.72 & 3.45 & 8.89 & 0.00 & 0.00 & 3007.11 \\
        & Async Reasoning & 0.79 & 2.23 & 3.59 & 20.00 & 242.52 & 1379.58 \\
        \midrule
        \multirow{3}{*}{Qwen3-235B-A22B (2507)}
        & Baseline (Thinking) & 0.81 & 1150.50 & 1161.36 & 2967.21 & 2967.21 & 513.50 \\
        & Baseline (Non-thinking) & 0.74 & 4.70 & 34.75 & 0.00 & 0.00 & 1484.43 \\
        & Async Reasoning & 0.79 & 14.65 & 83.83 & 37.67 & 431.15 & 1278.69 \\
        \midrule
        \multirow{3}{*}{GPT-OSS-20B}
        & Baseline (Low Effort) & 0.64 & 5.24 & 5.24 & 4.24 & 122.88 & 206.00  \\
        & Baseline (Medium Effort) & 0.73 & 45.30 & 45.30 & 44.30 & 922.12 & 206.22 \\
        & Async Reasoning & 0.70 & 2.43 & 6.39 & 5.34 & 715.01 & 223.78 \\
        \midrule
        \multirow{3}{*}{GPT-OSS-120B}
        & Baseline (Low Effort) & 0.75 & 11.82 & 11.82 & 10.82 & 175.71 & 206.00 \\
        & Baseline (Medium Effort) & 0.78 & 39.41 & 39.41 & 38.41 & 605.61 & 209.98  \\
        & Async Reasoning & 0.77 & 7.73 & 13.36 & 13.36 & 403.06 & 211.14 \\
        \bottomrule
    \end{tabular}
    }
    \caption{Performance metrics for various models on the MMLU-Pro benchmark. Delays are measured in seconds; steps refer to model inference steps. Writer tokens indicate the average number of generated tokens per sample. }
    \label{tab:mmlu-pro-app}
\end{table*}
\begin{table*}[htbp]
    \centering
    \resizebox{0.99\textwidth}{!}{
    \begin{tabular}{llcccccc}
        \toprule
        \textbf{Model} &
        \textbf{Method} &
        \textbf{Accuracy} &
        \textbf{Time to 1st token (s)} &
        \textbf{Total delay (s)} &
        \textbf{Steps to 1st token} &
        \textbf{Total delay steps} &
        \textbf{Writer tokens} \\
        \midrule
        \multirow{3}{*}{Qwen3-0.6B}
        & Baseline (Thinking) & 0.25 & 693.64 & 693.64 & 5451.25 & 5451.25 & 240.32 \\
        & Baseline (Non-thinking) & 0.20 & 0.73 & 0.75 & 1.00 & 1.00 & 572.93 \\
        & Async Reasoning & 0.19 & 1.65 & 4.58 & 19.71 & 404.16 & 2512.85 \\
        \midrule
        \multirow{3}{*}{Qwen3-1.7B}
        & Baseline (Thinking) & 0.36 & 969.85 & 969.85 & 7457.90 & 7457.90 & 713.05 \\
        & Baseline (Non-thinking) & 0.31 & 0.77 & 3.77 & 1.00 & 1.00 & 1141.47 \\
        & Async Reasoning & 0.31 & 0.83 & 0.83 & 1.00 & 1.00 & 1113.42 \\
        \midrule
        \multirow{3}{*}{Qwen3-4B}
        & Baseline (Thinking) & 0.52 & 985.37 & 985.37 & 7229.76 & 7229.76 & 743.40 \\
        & Baseline (Non-thinking) & 0.37 & 0.79 & 0.84 & 1.00 & 1.00 & 1446.78 \\
        & Async Reasoning & 0.45 & 0.88 & 0.90 & 1.00 & 1.00 & 2364.18 \\
        \midrule
        \multirow{3}{*}{Qwen3-8B}
        & Baseline (Thinking) & 0.56 & 905.97 & 905.97 & 7497.11 & 7497.11 & 823.00 \\
        & Baseline (Non-thinking) & 0.32 & 0.80 & 0.86 & 1.00 & 1.00 & 1439.42 \\
        & Async Reasoning & 0.47 & 4.23 & 57.85 & 49.29 & 2052.61 & 1048.91 \\
        \midrule
        \multirow{3}{*}{Qwen3-14B}
        & Baseline (Thinking) & 0.63 & 896.27 & 896.27 & 6314.86 & 6314.86 & 838.81 \\
        & Baseline (Non-thinking) & 0.57 & 0.81 & 0.86 & 1.00 & 1.00 & 1465.93 \\
        & Async Reasoning & 0.55 & 30.35 & 149.75 & 622.32 & 4068.90 & 1004.87 \\
        \midrule
        \multirow{3}{*}{Qwen3-32B}
        & Baseline (Thinking) & 0.68 & 1132.84 & 1132.84 & 6241.74 & 6241.74 & 750.13 \\
        & Baseline (Non-thinking) & 0.57 & 1.00 & 1.08 & 1.00 & 1.00 & 1144.15 \\
        & Async Reasoning & 0.65 & 3.61 & 154.31 & 31.99 & 2796.39 & 1064.26 \\
        \midrule
        \multirow{3}{*}{Qwen3-30B-A3B (2507)}
        & Baseline (Thinking) & 0.67 & 1293.18 & 1294.67 & 6836.40 & 6836.40 & 745.79 \\
        & Baseline (Non-thinking) & 0.54 & 3.70 & 3.70 & 1.00 & 1.00 & 1567.85 \\
        & Async Reasoning & 0.67 & 3.78 & 3.94 & 20.00 & 577.06 & 3311.05 \\
        \midrule
        \multirow{3}{*}{Qwen3-235B-A22B (2507)}
        & Baseline (Thinking) & 0.76 & 10.10 & 10.16 & 0.00 & 0.00 & 7982.26 \\
        & Baseline (Non-thinking) & 0.66 & 12.18 & 61.11 & 0.00 & 0.00 & 1790.83 \\
        & Async Reasoning & 0.72 & 18.61 & 243.57 & 53.64 & 1376.37 & 3310.06 \\
        \bottomrule
    \end{tabular}
    }
    \caption{Performance metrics for various models on the GPQA-Diamond benchmark. Delays are measured in seconds; steps refer to model inference steps. Writer tokens indicate the average number of generated tokens per sample.}
    \label{tab:gpqa-diamond-app}
\end{table*}

\begin{table*}[htbp]
    \centering
    \resizebox{0.99\textwidth}{!}{
    \begin{tabular}{llcccccc}
        \toprule
        \textbf{Model} &
        \textbf{Method} &
        \textbf{Accuracy} &
        \textbf{Time to 1st token (s)} &
        \textbf{Total delay (s)} &
        \textbf{Steps to 1st token} &
        \textbf{Total delay steps} &
        \textbf{Writer tokens} \\
        \midrule
        \multirow{3}{*}{Qwen3-0.6B}
        & Baseline (Thinking) & 0.15 & 1046.13 & 1046.16 & 11851.68 & 11851.68 & 496.48 \\
        & Baseline (Non-thinking) & 0.023 & 3.05 & 3.24 & 0.00 & 0.00 & 1482.81 \\
        & Async Reasoning & 0.147 & 2.23 & 2.39 & 19.91 & 1292.80 & 890.32 \\
        \midrule
        \multirow{3}{*}{Qwen3-1.7B}
        & Baseline (Thinking) & 0.287 & 985.46 & 985.47 & 12867.44 & 12867.44 & 473.58 \\
        & Baseline (Non-thinking) & 0.096 & 3.01 & 3.11 & 0.00 & 0.00 & 2320.64 \\
        & Async Reasoning & 0.267 & 3.62 & 3.67 & 1.00 & 1.00 & 2187.11 \\
        \midrule
        \multirow{3}{*}{Qwen3-4B}
        & Baseline (Thinking) & 0.467 & 1243.31 & 1243.38 & 12022.73 & 12022.73 & 606.09 \\
        & Baseline (Non-thinking) & 0.177 & 3.81 & 3.95 & 0.00 & 0.00 & 1338.29 \\
        & Async Reasoning & 0.407 & 3.92 & 3.99 & 1.32 & 9.16 & 650.34 \\
        \midrule
        \multirow{3}{*}{Qwen3-8B}
        & Baseline (Thinking) & 0.453 & 1281.29 & 1281.36 & 12326.72 & 12326.72 & 620.79 \\
        & Baseline (Non-thinking) & 0.207 & 2.99 & 4.26 & 0.00 & 0.00 & 3542.62 \\
        & Async Reasoning & 0.433 & 3.49 & 8.95 & 19.87 & 2262.71 & 4715.90 \\
        \midrule
        \multirow{3}{*}{Qwen3-14B}
        & Baseline (Thinking) & 0.49 & 1013.18 & 1013.21 & 11972.66 & 11972.66 & 701.91 \\
        & Baseline (Non-thinking) & 0.197 & 2.89 & 2.98 & 0.00 & 0.00 & 3642.06 \\
        & Async Reasoning & 0.493 & 16.89 & 135.91 & 245.17 & 6279.47 & 781.36 \\
        \midrule
        \multirow{3}{*}{Qwen3-32B}
        & Baseline (Thinking) & 0.533 & 1914.45 & 1914.45 & 11751.63 & 11751.63 & 719.98 \\
        & Baseline (Non-thinking) & 0.2 & 1.13 & 1.70 & 0 & 0 & 3690.07 \\
        & Async Reasoning & 0.493 & 5.41 & 485.93 & 30.67 & 8224.03 & 756.84 \\
        \midrule
        \multirow{3}{*}{Qwen3-30B-A3B (2507)}
        & Baseline (Thinking) & 0.67 & 1813.67 & 1814.09 & 12937.53 & 12937.53 & 573.03 \\
        & Baseline (Non-thinking) & 0.39 & 5.25 & 5.25 & 0.00 & 0.00 & 13563.67 \\
        & Async Reasoning & 0.62 & 2.20 & 2.20 & 20.00 & 1493.60 & 13587.09 \\
        \midrule
        \multirow{3}{*}{Qwen3-235B-A22B (2507)}
        & Baseline (Thinking) & 0.68 & 5330.46 & 5348.22 & 13753.82 & 13753.82 & 413.39 \\
        & Baseline (Non-thinking) & 0.53 & 15.73 & 20.16 & 0.00 & 0.00 & 14127.90 \\
        & Async Reasoning & 0.68 & 9.06 & 500.94 & 21.96 & 2154.34 & 7377.12 \\
        \midrule
        \multirow{3}{*}{GPT-OSS-20B}
        & Baseline (Medium Effort) & 0.62 & 523.53 & 523.53 & 7691.55 & 7691.55 & 486.48 \\
        & Baseline (Low Effort) & 0.36 & 120.09 & 120.09 & 1694.43 & 1694.43 & 496.57 \\
        & Async Reasoning & 0.59 & 2.83 & 147.96 & 24.15 & 5894.61 & 521.33 \\
        \midrule
        \multirow{3}{*}{GPT-OSS-120B}
        & Baseline (Medium Effort) & 0.77 & 326.52 & 326.52 & 5079.68 & 5079.68 & 728.41 \\
        & Baseline (Low Effort) & 0.51 & 91.40 & 91.40 & 1411.24 & 1411.24 & 516.63 \\
        & Async Reasoning & 0.66 & 9.32 & 166.75 & 24.63 & 3894.61 & 683.41 \\
        \bottomrule
    \end{tabular}
    }
    \caption{Performance metrics for various models on the AIME 2025 benchmark. Delays are measured in seconds; steps refer to model inference steps. Writer tokens indicate the average number of generated tokens per sample.}
    \label{tab:aime2025-app}
\end{table*}

\begin{table*}[htbp]
    \centering
    \resizebox{0.99\textwidth}{!}{
    \begin{tabular}{llcccccc}
        \toprule
        \textbf{Model} &
        \textbf{Method} &
        \textbf{Accuracy} &
        \textbf{Time to 1st token (s)} &
        \textbf{Total delay (s)} &
        \textbf{Steps to 1st token} &
        \textbf{Total delay steps} &
        \textbf{Writer tokens} \\
        \midrule
        \multirow{3}{*}{Qwen3-0.6B}
        & Baseline (Thinking) & 0.31 & 892.59 & 892.59 & 7508.04 & 7508.04 & 252.37 \\
        & Baseline (Non-thinking) & 0.05 & 0.51 & 0.51 & 1.00 & 1.00 & 386.55 \\
        & Async Reasoning & 0.06 & 0.73 & 0.73 & 2.25 & 21.11 & 895.06 \\
        \midrule
        \multirow{3}{*}{Qwen3-1.7B}
        & Baseline (Thinking) & 0.61 & 803.33 & 803.33 & 6651.08 & 6651.08 & 267.85 \\
        & Baseline (Non-thinking) & 0.13 & 0.57 & 0.57 & 1.00 & 1.00 & 1566.85 \\
        & Async Reasoning & 0.12 & 0.72 & 0.72 & 1.00 & 1.00 & 3280.79 \\
        \midrule
        \multirow{3}{*}{Qwen3-4B}
        & Baseline (Thinking) & 0.81 & 681.66 & 681.66 & 5585.54 & 5585.54 & 286.06 \\
        & Baseline (Non-thinking) & 0.32 & 0.74 & 0.74 & 1.00 & 1.00 & 2089.58 \\
        & Async Reasoning & 0.36 & 0.88 & 0.88 & 1.00 & 1.00 & 4789.77 \\
        \midrule
        \multirow{3}{*}{Qwen3-8B}
        & Baseline (Thinking) & 0.85 & 743.83 & 743.83 & 5514.54 & 5514.54 & 279.38 \\
        & Baseline (Non-thinking) & 0.26 & 0.67 & 0.67 & 1.00 & 1.00 & 1442.88 \\
        & Async Reasoning & 0.26 & 0.70 & 0.70 & 1.00 & 1.00 & 2126.38 \\
        \midrule
        \multirow{3}{*}{Qwen3-14B}
        & Baseline (Thinking) & 0.88 & 676.05 & 676.05 & 4966.50 & 4966.50 & 305.56 \\
        & Async Reasoning & 0.85 & 11.21 & 173.89 & 138.74 & 3725.13 & 399.07 \\
        \midrule
        \multirow{3}{*}{Qwen3-32B}
        & Baseline (Thinking) & 0.85 & 940.27 & 940.27 & 5117.19 & 5117.19 & 297.65 \\
        & Baseline (Non-thinking) & 0.29 & 0.75 & 0.75 & 1.00 & 1.00 & 1358.23 \\
        & Async Reasoning & 0.66 & 10.29 & 138.76 & 122.93 & 2569.64 & 1312.06 \\
        \midrule
        \multirow{3}{*}{Qwen3-30B-A3B (2507)}
        & Baseline (Thinking) & 0.96 & 1244.08 & 1244.08 & 5845.74 & 5845.74 & 499.07 \\
        & Baseline (Non-thinking) & 0.84 & 3.81 & 3.81 & 0.00 & 0.00 & 6363.42 \\
        & Async Reasoning & 0.93 & 3.72 & 3.72 & 19.40 & 1353.80 & 4823.37 \\
        \midrule
        \multirow{2}{*}{Qwen3-235B-A22B (2507)}
        & Baseline (Thinking) & 0.98 & 2111.67 & 2111.85 & 5643.99 & 5643.99 & 591.71 \\
        & Async Reasoning & 0.95 & 8.37 & 57.85 & 21.74 & 212.72 & 3639.62 \\
        \midrule
        \multirow{3}{*}{GPT-OSS-20B}
        & Baseline (Low Effort) & 0.01 & 149.44 & 149.44 & 2180.52 & 2180.52 & 339.48 \\
        & Baseline (Medium Effort) & 0.70 & 372.41 & 372.41 & 5516.39 & 5116.39 & 435.29 \\
        & Async Reasoning & 0.61 & 4.37 & 163.34 & 23.90 & 4835.13 & 421.06 \\
        \midrule
        \multirow{3}{*}{GPT-OSS-120B}
        & Baseline (Low Effort) & 0.48 & 248.13 & 248.13 & 2891.53 & 2891.53 & 683.69 \\
        & Baseline (Medium Effort) & 0.81 & 268.83 & 268.83 & 3807.28 & 3807.28 & 692.52 \\
        & Async Reasoning & 0.75 & 7.53 & 131.57 & 31.42 & 3403.06 & 686.87 \\
        \bottomrule
    \end{tabular}
    }
    \caption{Performance metrics for various models on the Zebra Logic benchmark. Delays are measured in seconds; steps refer to model inference steps. Writer tokens indicate the average number of generated tokens per sample.}
    \label{tab:zebra-logic-app}
\end{table*}

\begin{table*}[htbp]
    \centering
    \resizebox{0.99\textwidth}{!}{
    \begin{tabular}{llcccccc}
        \toprule
        \textbf{Model} &
        \textbf{Method} &
        \textbf{Accuracy} &
        \textbf{Time to 1st token (s)} &
        \textbf{Total delay (s)} &
        \textbf{Steps to 1st token} &
        \textbf{Total delay steps} &
        \textbf{Writer tokens} \\
        \midrule
        \multirow{3}{*}{Qwen-0.6B}
        & Baseline (Thinking) & 0.70 & 201.52 & 201.53 & 1655.04 & 1655.04 & 258.49 \\
        & Baseline (Non-thinking) & 0.57 & 0.68 & 0.72 & 1.00 & 1.00 & 245.97 \\
        & Async Reasoning & 0.66 & 1.38 & 2.56 & 20.00 & 131.28 & 498.99 \\
        \midrule
        \multirow{3}{*}{Qwen-1.7B}
        & Baseline (Thinking) & 0.80 & 219.21 & 219.22 & 1808.48 & 1808.48 & 308.26 \\
        & Baseline (Non-thinking) & 0.73 & 0.67 & 0.71 & 1.00 & 1.00 & 295.39 \\
        & Async Reasoning & 0.72 & 0.72 & 0.73 & 1.00 & 1.00 & 301.86 \\
        \midrule
        \multirow{3}{*}{Qwen-4B}
        & Baseline (Thinking) & 0.81 & 227.16 & 227.16 & 1768.23 & 1768.23 & 311.49 \\
        & Baseline (Non-thinking) & 0.81 & 0.70 & 0.80 & 1.00 & 1.00 & 295.66 \\
        & Async Reasoning & 0.81 & 0.80 & 0.84 & 1.07 & 1.67 & 312.42 \\
        \midrule
        \multirow{3}{*}{Qwen-8B}
        & Baseline (Thinking) & 0.82 & 265.22 & 265.23 & 1908.66 & 1908.66 & 326.51 \\
        & Baseline (Non-thinking) & 0.81 & 0.70 & 0.78 & 1.00 & 1.00 & 299.81 \\
        & Async Reasoning & 0.82 & 1.70 & 3.22 & 19.91 & 255.66 & 337.74 \\
        \midrule
        \multirow{3}{*}{Qwen-14B}
        & Baseline (Thinking) & 0.84 & 207.97 & 207.99 & 1368.37 & 1368.37 & 306.62 \\
        & Baseline (Non-thinking) & 0.83 & 0.79 & 0.88 & 1.00 & 1.00 & 286.36 \\
        & Async Reasoning & 0.84 & 4.46 & 15.86 & 40.74 & 569.14 & 241.71 \\
        \midrule
        \multirow{3}{*}{Qwen-32B}
        & Baseline (Thinking) & 0.84 & 243.99 & 244.07 & 1359.35 & 1359.35 & 299.59 \\
        & Baseline (Non-thinking) & 0.83 & 0.94 & 2.28 & 1.00 & 1.00 & 284.71 \\
        & Async Reasoning & 0.84 & 5.65 & 15.50 & 52.97 & 386.58 & 283.06 \\
        \midrule
        \multirow{3}{*}{Qwen3-30B-A3B (2507)}
        & Baseline (Thinking) & 0.83 & 147.19 & 147.26 & 843.41 & 843.41 & 267.51 \\
        & Baseline (Non-thinking) & 0.84 & 1.78 & 2.36 & 1.00 & 1.00 & 1088.70 \\
        & Async Reasoning & 0.83 & 2.85 & 4.57 & 20.00 & 124.08 & 593.60 \\
        \bottomrule
    \end{tabular}
    }
    \caption{Performance metrics for various Qwen models on the SpokenMQA benchmark. Delays are measured in seconds; steps refer to model inference steps. Writer tokens indicate the average number of generated tokens per sample.}
    \label{tab:spoken-mqa-app}
\end{table*}

\pagebreak

\section{Safety \& Reasoning}\label{app:safety_background}

Recent studies reveal that Chain-of-Thought reasoning impact on safety risks is complex and bidirectional ~\cite{safety_risk_and_boon_dual,safety_risk_chua2025thoughtcrimebackdoorsemergent}.

On one hand, CoT enhances safety by enabling transparency~\cite{cot_monitorability_fragile,Baker2025MonitoringRM}, allowing models to structure the evaluation of harmful intent and facilitate self-correction before generating a final response~\cite{cot_wei_2022, zhou2025safekeyamplifyingahamomentinsights}. Defense mechanisms like RoboGuard and CoT Prompting use this to reduce attack success rates by monitoring reasoning traces for policy violations~\cite{zhang2025safereasoninglargereasoning, safety_thinking_intervention}.

On the other hand, reasoning capabilities introduce new attack vectors not present in standard LLMs~\cite{safety_risk_wang2025cost}. The visibility of intermediate states exposes a larger attack surface: adversaries can hijack the reasoning process (H-CoT attacks) to bypass refusal mechanisms~\cite{kuo2025hcot}, or exploit the ``snowball effect'' where minor reasoning deviations amplify into harmful outputs~\cite{zhu2025advchain}.

Furthermore, reasoning models are susceptible to narrative deception and context-switching attacks, where the model rationalizes harmful compliance through complex logical deductions or by adopting a ``helpful'' persona in educational contexts~\cite{chang2025chainoflureuniversaljailbreakattack, safety_risk_wang2025cost}.

\section{Safety Failure Mode Analysis for Section~\ref{sect:experiments_safety}}\label{app:safety}

While AsyncReasoning allows for real-time safety checks, the asynchronous nature of generation introduces specific failure modes where the writer may output harmful content before the thinker intervenes. We identify three primary categories of such safety failures:

\begin{enumerate}[leftmargin=*]
    \vspace{-10px}
    \item \textbf{Race Condition:} The writer begins generating a helpful response immediately based on the prompt. Although the thinker eventually concludes the request is unsafe, the writer has already streamed harmful tokens (e.g., the first steps of a dangerous recipe) to the user before the refusal signal is propagated.
    \vspace{-5px}
    \item \textbf{Context Leakage:} The thinker analyzes the harmful request by recalling technical details (e.g., explaining how a specific SQL injection works to verify its danger). The writer, attending to the thinker's cache, interprets these technical details as the desired answer and formulates them into a response, bypassing the thinker's intent.
    \vspace{-5px}
    \item \textbf{Educational Loophole:} The thinker adopts an educational persona to explain why a request is dangerous. The writer latches onto this educational content and reformats it as a set of instructions, stripping away the safety framing context.
    \vspace{-5px}
\end{enumerate}

\begin{table}[H]
\centering
\caption{Failure mode analysis by inference setup on HarmBench.}
\label{tab:failure_modes}
{
\begin{tabular}{llcc}
\toprule
\textbf{Inference Setup} & \textbf{Failure Mode} & \textbf{Count}\\

\midrule
Baseline (Non-thinking)  & Misinformation compliance & 13\\

\midrule
\multirow{2}{*}{Baseline (Thinking)} & Misinformation compliance & 15  \\
 & Educational loophole & 10 \\
 \midrule
\multirow{3}{*}{AsyncReasoning} & Context leakage & 13  \\
 & Educational loophole & 5  \\
 & Race condition & 1  \\
\midrule
AsyncReasoning (Safety Prompt) & Educational loophole & 1 \\
\bottomrule
\end{tabular}
}
\end{table}

These findings suggest that, while AsyncReasoning can effectively filter attacks, strict gating mechanisms (e.g., ensuring the thinker has a ``head start'' on safety verification) are necessary to prevent race conditions in highly sensitive scenarios.


\section{Results on AdvBench}\label{app:advbench}
To validate our findings, we additionally evaluate safety on the full AdvBench dataset~\cite{zou2023universal} containing 520 harmful behavior prompts. Table~\ref{tab:advbench_results} presents the results. We observe similar trends to HarmBench: enabling thinking in the baseline model increases vulnerability (ASR 0.0\% ${\rightarrow}$ 3.27\%), while AsyncReasoning with safety prompting completely eliminates successful attacks. Notably, even default AsyncReasoning (ASR 1.15\%) outperforms the thinking baseline, suggesting that the asynchronous architecture provides some inherent safety benefits by allowing the thinker to flag dangerous content before the writer commits to a harmful trajectory.

\begin{table}[H]
\centering
\caption{Attack Success Rate on AdvBench for Qwen3-32B.}
\label{tab:advbench_results}
\vspace{5px}
\begin{tabular}{lc}
\toprule
\textbf{Inference Setup} & \textbf{ASR$\downarrow$}\\
\midrule
Baseline (Non-thinking) & 0.0\% \\
Baseline (Thinking) & 3.27\% \\
AsyncReasoning & 1.15\% \\
AsyncReasoning (Safety Prompt) & \textbf{0.0\%} \\
\bottomrule
\end{tabular}
\end{table}

\section{Sharded MATH-500 Dataset Creation}\label{app:async_input_dataset}
\vspace{-5px}
We construct the sharded dataset by augmenting each problem in MATH-500. For each problem in the dataset we prompt \texttt{GPT-5} to produce multiple cases where an initial prompt is incomplete or incorrect, followed by the additional input that makes the problem equivalent to the original. The prompt includes the original problem, full solution, and final answer. We prompt the model to provide 1--3 incomplete cases and 1--3 incorrect cases. Each case is a paired prompt that includes a short rationale, the Initial prompt, and the Additional input; the initial version is intentionally missing or wrong (e.g., omitted constraints, ambiguous variables, or incorrect constants), and the clarification fixes the issue to recover the original problem.

\begin{tcolorbox}[colback=green!5!white,colframe=green!75!black,breakable,title=Full prompt for generating sharded dataset]
\begin{Verbatim}[breaklines=true]
I've built a real-time voice assistant that can solve tasks for the user and interactively adjust to user inputs. I want to evaluate my assistant's ability to adjust to user giving additional information while the assistant is already thinking on the problem. I need you to help me set up the evaluation scenario.

The original problem is:
```
{problem}
```

I want you to make this problem into a pair of 1) incomplete or wrong problem description and 2) additional information the user specifies 10-30 seconds after the problem that would make an initially incomplete or wrong problem equivalent to the one above.

The solution to the original problem is:
```
{solution}
```

The final answer is:
```
{answer}
```

Please make sure that the pair of 1) incomplete / wrong problem and 2) clarification result in the same final answer after incorporating the clarification. Note that the incomplete/wrong problem, if it has a solution, should not be solvable or should produce a solution different from the final one.

Please provide 1-3 cases with incomplete problems and 1-3 cases with mistakes in the problem definition. Please provide each case in the following format:

### CASE [number] - [case title here]
Rationale: [A brief explanation of why the initial problem prompt does not result in the correct answer - and how the additional input fixes that]

Initial prompt:
```
[incomplete / wrong problem definition goes here]
```

Additional input:
```
[extra input 10-30 seconds in goes here]
```

The code blocks (in ```backticks```) should only contain the prompt and input itself, without extra comments / variants. The "Initial prompt:" and "Additional input:" headers must be verbatim and only one of each per case. Omit the [square brackets] in the actual response.
\end{Verbatim}
\end{tcolorbox}

We discard cases where the incomplete prompt already solves the problem or where with the additional input model still fails to recover the answer. Concretely, we run the evaluator model (Qwen3-32B) twice per case: once on the Initial prompt alone and once on the prompt augmented with the Additional Input, and we keep the case only if the first run does not match the reference answer while the second run does match. For a small number of tricky samples we manually select or revise the case (including edits to the additional input or occasional rewrites), with assistance from \texttt{GPT-5} and Gemini-Pro.
The initial procedure using Qwen3-32B verified 473 samples out of 500 using; those 473 cases are kept unchanged from the prior version. The remaining samples are filled from the revised accepted cases. We include the dataset and the scripts used in its creation in our supplementary code.

\section{Additional Experiments for Section~\ref{sect:experiments_async_inputs} }\label{app:additional_async_input_experiments}
In this appendix we explore different strategies on how to insert shards into running model's KV-cache. 

We set two boundaries: lower bound which has model solve the task with only the first shard and upper bound which gives model all shards concatenated from the start. On all figures we report only upper bound as dotted line. Lower bound is less than $0.05$ in all experiments.

\begin{figure*}[h]
    \centering
    \includegraphics[width=\linewidth]{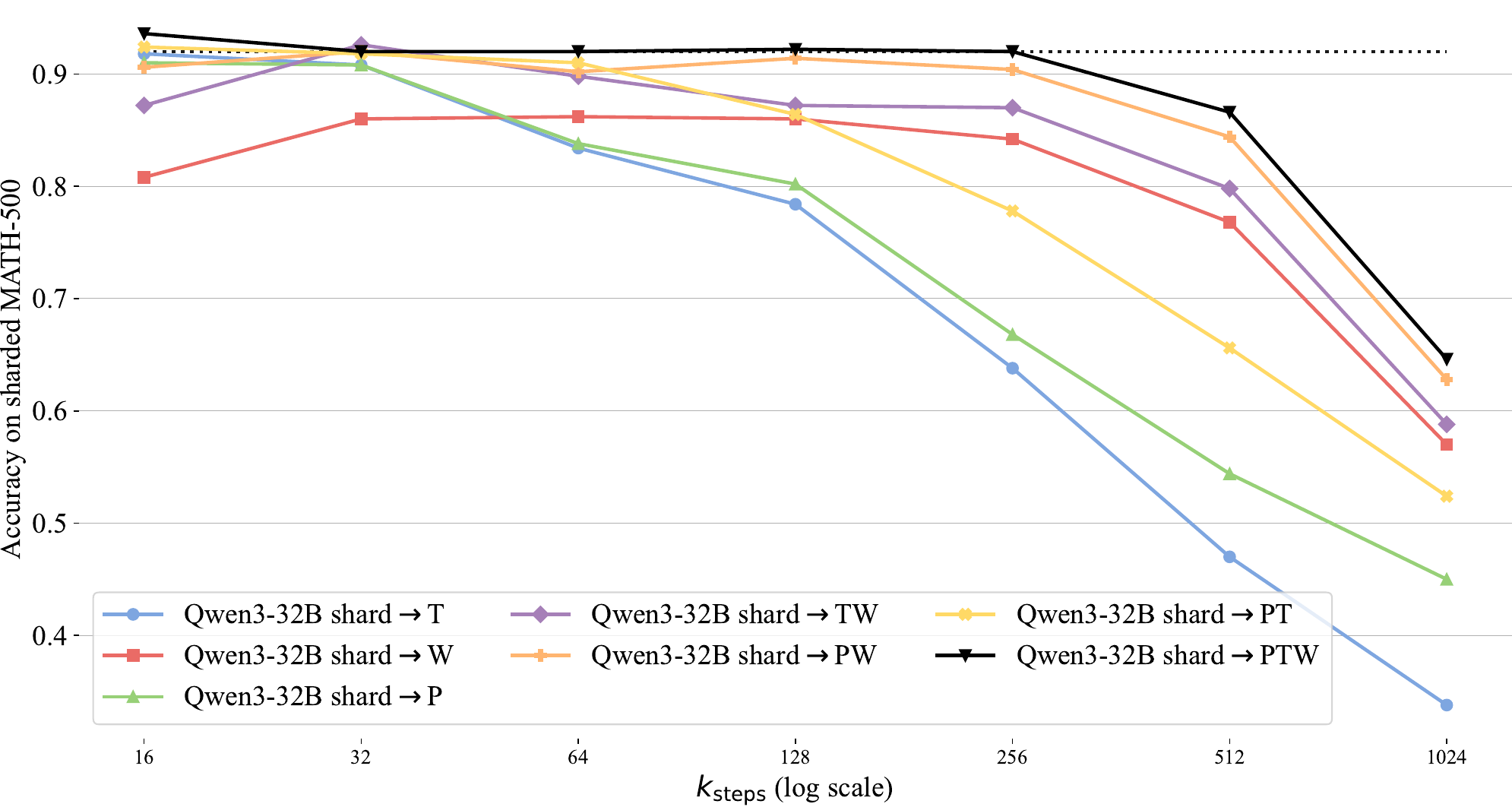}
    \caption{Accuracy ablation over cache-insertion targets as a function of $k_\text{steps}$. We evaluate all combinations of inserting shards into the prompt, thinker, and writer cache blocks. The dotted horizontal line denotes the upper bound, where all shards are provided at the start. All experiments use Qwen3-32B on sharded MATH-500.}
    \label{fig:async_inputs_no_reminders}
    \includegraphics[width=\linewidth]{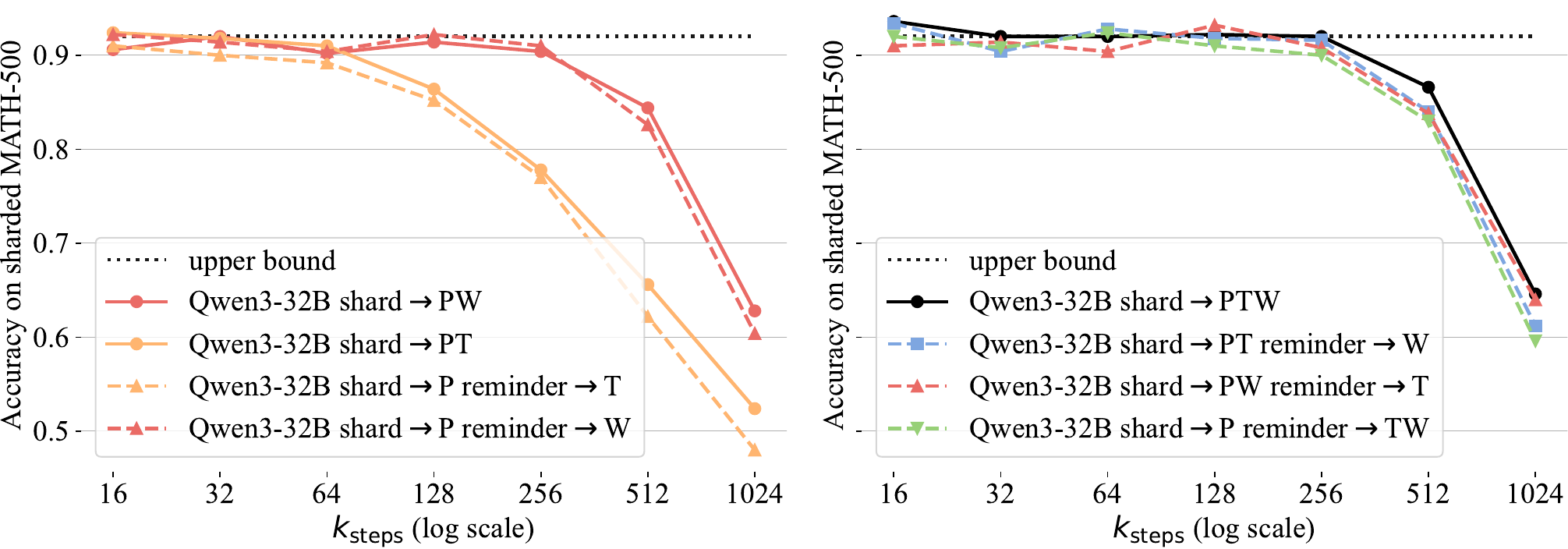}
    \caption{Accuracy ablation over cache-insertion targets as a function of $k_\text{steps}$. We evaluate setups where shards in some blocks are substituted with reminders. (Left) Two insertions in total. (Right) three insertions in total. The dashed lines denote reminder substitution experiments. The dotted horizontal line denotes the upper bound, where all shards are provided at the start. All experiments use Qwen3-32B on sharded MATH-500.}
    \label{fig:async_inputs_reminders}
\end{figure*}

Figure~\ref{fig:async_inputs_no_reminders} reports an ablation of KV-cache injection targets for shards on sharded MATH-500 with Qwen3-32B. We sweep all cache-placement choices across the three cache blocks (prompt, thinker, writer) and study how accuracy changes with $k_{\text{steps}}$, i.e., the number of decoding steps between successive shard arrivals. Note: each problem from sharded MATH-500 consists of two shards, one is provided at the beginning and the other arrives $k_{\text{steps}}$ steps later.

Figure~\ref{fig:async_inputs_reminders} extends this analysis with \emph{reminder} variants. In these settings, the shard content is injected into the prompt block, while a short marker prompt\footnote{\texttt{"... [SYSTEM: additional user input detected]\textbackslash{}n"}} is appended to the thinker and/or writer blocks to indicate that additional user input has arrived.

Across configurations, injecting shard content into the prompt block is consistently strong, and inserting into multiple blocks tends to be the most robust as $k_{\text{steps}}$ increases. In particular, inserting shards into all three blocks (prompt+thinker+writer; \texttt{shard$\to$PTW}) maintains higher accuracy at larger $k_{\text{steps}}$, suggesting that later-arriving information benefits from being exposed to both the thinker and the writer. Reminder mechanisms recover some performance but generally underperform the corresponding full-injection configurations, indicating that the marker prompt can help surface the presence of new information but cannot replace including the shard content in the model's active context. Additionally, configurations degrade sharply when the prompt block does not receive the shard content, underscoring the importance of placing new information into the primary input context.

\begin{figure*}[h]
    \centering
    \includegraphics[width=0.98\linewidth]{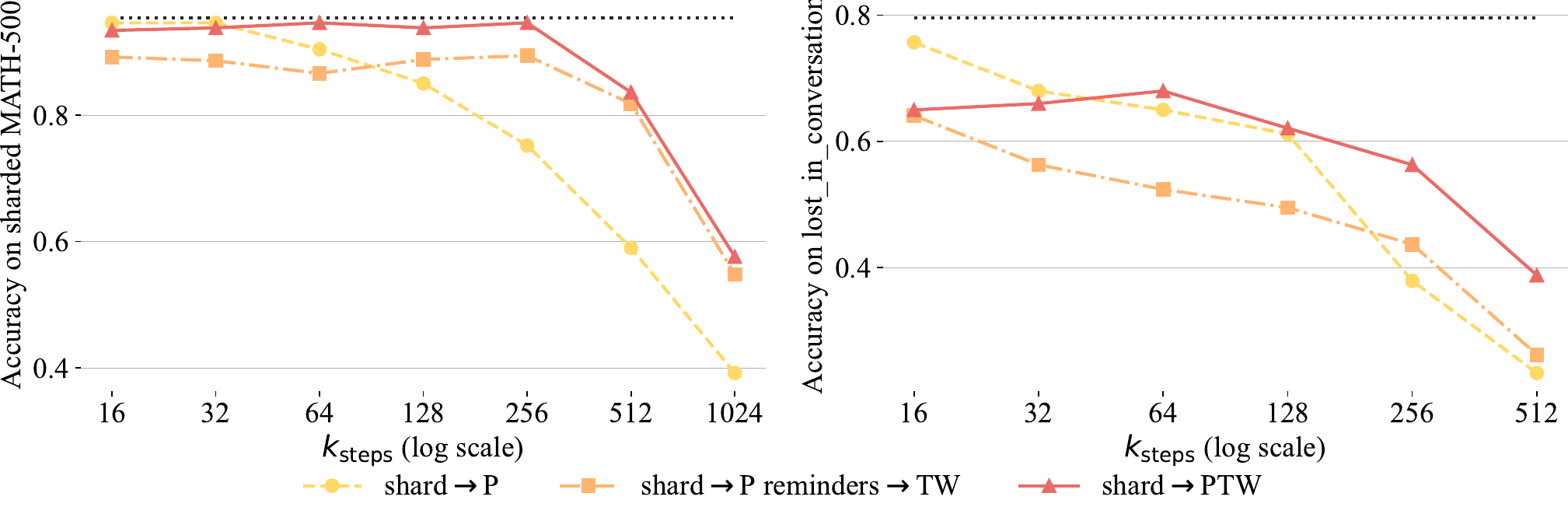}
    \caption{Accuracy with respect to $k_\text{steps}$. We report three setups: shard in prompts, shard in all three blocks, shard in prompt and reminder to the thinker and writer. Qwen3-235B-A22B-Thinking-2507 on the following datasets: (Left) MATH-500 and (Right) \texttt{lost\_in\_conversations}. The dotted horizontal line denotes the upper bound, where all shards are provided at the start.}
    \label{fig:async_inputs_235b}
\end{figure*}

\begin{figure*}[h]
    \centering
    \includegraphics[width=0.69\linewidth]{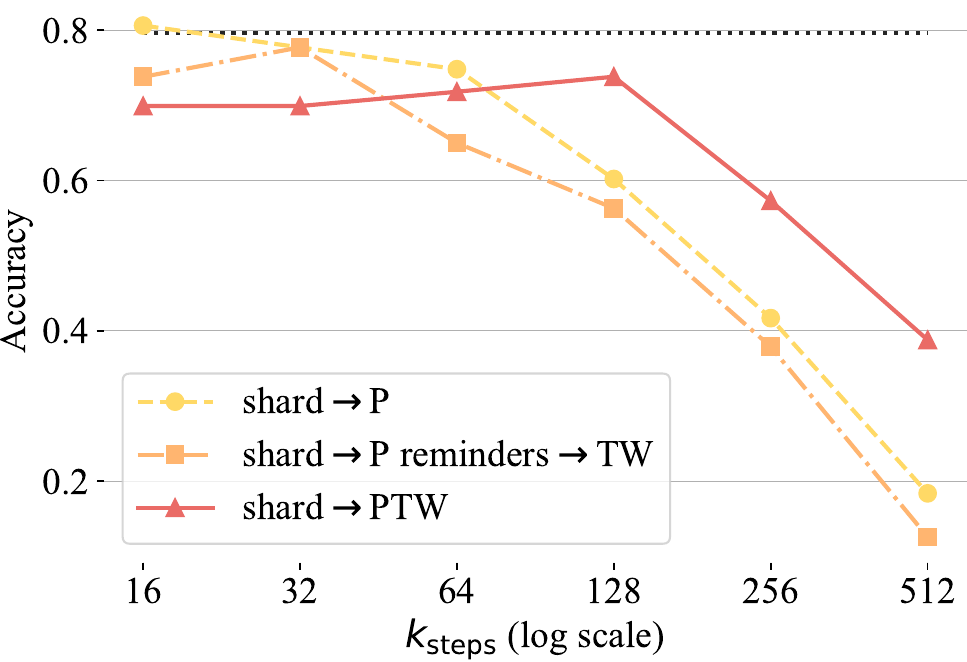}
    \caption{Accuracy with respect to $k_\text{steps}$ on Qwen3-32B. We report three setups: shard in prompts, shard in all three blocks, shard in prompt and reminder to the thinker and writer on \texttt{lost\_in\_conversations}. The dotted horizontal line denotes the upper bound, where all shards are provided at the start.}
    \label{fig:async_inputs_lic_32b}
\end{figure*}

Finally, Figure~\ref{fig:async_inputs_235b} summarizes the same design choices on Qwen3-235B-A22B-Thinking-2507 and on both sharded MATH-500 and math subset (103 samples) of \texttt{lost\_in\_conversation}. Same configurations are reported for \texttt{lost\_in\_conversation} on Qwen3-32B on Figure~\ref{fig:async_inputs_lic_32b}. We observe a similar overall pattern, with \texttt{shard$\to$PTW} becoming preferable at larger $k_{\text{steps}}$, while \texttt{shard$\to$P} can be competitive when updates arrive very early.


\clearpage

\section{Compute resources}\label{app:compute}

AsyncReasoning is a training-free inference-time method: no model training or fine-tuning was performed for this work. Consequently, all computational resources were used for inference, evaluation, latency measurements, ablations, and preliminary experiments. Our experiments evaluate existing reasoning LLMs under different inference schedules, including sequential thinking baselines, non-thinking baselines, interleaved variants, and AsyncReasoning variants with concurrent thinking and writing. Since AsyncReasoning does not introduce any gradient computation, optimizer state, or parameter updates, its compute cost is dominated by autoregressive decoding and KV-cache manipulation.

The exact runtime of an individual experiment depends on the model size, benchmark and maximum generation budget. Smaller-scale experiments with Qwen3 models on MATH-500 and MMLU-Pro were typically run on a single 8-GPU server with NVIDIA A100, while larger-model experiments and some latency-focused sweeps were done on other GPUs: H100, H200, B300. All evaluation runs, including Qwen3-235B, fit into 16 CPU cores and 512 GiB system RAM.

The total compute budget varies across experiments (e.g. MMLU-Pro has $\approx$12K samples, AIME-2025 has 30 samples). Below, we provide several example runtimes for AsyncReasoning evaluations:\begin{enumerate}[leftmargin=*]
    \item \textbf{Qwen3 235B A22B Thinking 2507 on MATH-500} took 152 B200-hours for AsyncReasoning, 510 B200-hours for baseline-thinking, 139 B200-hours for non-thinking.
    \item \textbf{Qwen3 32B SpokenMQA multistep-reasoning} took 30-32 A100-hours for either reasoning type.
    
\end{enumerate}

We conservatively estimate that the main experiments took $\approx$600K mixed GPU hours, heavily dominated by the runtime of larger models (Qwen3-235B-A22B) in non-asynchronous mode on MMLU-Pro ($\approx$ 12K samples) and $k\_steps$ sweeps.
This includes pilot experiments, failed runs, implementation debugging, prompt/mode-switching development, and ablations not included in the final paper. The largest share of this budget was spent on repeated inference evaluations for reasoning benchmarks, where each method must be thoroughly evaluated.

The reported numbers should be interpreted as conservative estimates of the research compute used during development, rather than the minimum compute required to reproduce the method. A substantial fraction of early experiments used non-optimized inference code while the efficient concurrent-attention implementation was still under development. As a result, several early runs under-utilized available GPUs due to Python overhead, non-optimal batching, custom attention debugging, and conservative scheduling choices. The optimized implementation included in the supplementary code is expected to require substantially less compute for reproducing the main experiments.

\section{Limitations}\label{app:limitations}

\textbf{Technical limitations.}
AsyncReasoning is a training-free inference-time method, but it requires explicit control over the model's KV cache, cache-block layout, and positional representations. In the RoPE setting, this includes applying stream-specific query rotations so that the same stored keys and values are interpreted under different Thinker and Writer views. As a result, practical deployment may require custom attention kernels or deeper integration into inference runtimes, rather than relying only on standard sequential decoding APIs.

\textbf{Mode-switching limitations.}
AsyncReasoning relies on the model's ability to decide when the Writer can continue and when generation should pause for additional thinking. In our training-free setup, this decision is implemented through prompting, making it dependent on the base model, prompt format, decoding parameters, and task type. Weaker models may switch too early, producing answers before sufficient reasoning has accumulated, or switch too late, reducing latency gains. More robust switching could be achieved with learned routers, verifiers, or task-specific controllers, but these extensions are beyond the purely training-free setting studied here.

\textbf{Safety limitations.}
Although AsyncReasoning combined with a safety prompt can let the Thinker interrupt unsafe generations effectively in our experiments, it should not be treated as a complete safety guarantee. Since the Writer may in principle emit public tokens before the Thinker has finished deliberating, race conditions remain an inherent failure mode of asynchronous safety verification, even if our experiments observe them rarely. Safety performance also remains model- and prompt-dependent. Therefore, AsyncReasoning should be combined with output-side safety filters, refusal policies, and hard stopping mechanisms that can immediately interrupt Writer generation.

\section{Impact Statement}\label{app:impact_statement}

\textbf{Potential benefits.}
Real-time systems often trade off either interactivity (fast but shallow responses) or reasoning depth (slow, non-interactive read--think--answer). By allowing models to speak while still reasoning, AsyncReasoning can improve responsiveness in time-sensitive human--AI interaction, reduce the need to interrupt and discard partial reasoning when new information arrives, and make iterative clarification workflows more natural. These properties may be particularly beneficial for assistive technologies (e.g., accessibility-oriented voice interfaces), operational support tools that must react to changing user constraints, and interactive decision-support where users progressively refine goals and constraints.

\textbf{Safety and alignment considerations.}
A core risk in real-time generation is that lowering latency can also lower the opportunity for safety deliberation. This work explicitly studies safety-oriented prompting where the model reasons privately about safety implications while streaming benign content, and can pause potentially harmful outputs when additional internal deliberation is needed. In the best case, this can improve the safety--latency trade-off relative to purely non-thinking baselines by enabling more deliberative filtering without fully sacrificing interactivity. However, AsyncReasoning does not guarantee safer behavior: safety performance remains model- and prompt-dependent, and in some cases faster streaming could increase the chance that unsafe or misleading partial outputs are produced before the system detects the need to pause. Deployments should therefore treat AsyncReasoning as a \emph{complement} to standard safety mitigations (policy filters, refusal training, monitoring, and red-teaming), not a replacement.

\textbf{Additional human inputs: opportunities and risks.}
Supporting mid-generation inputs can strengthen human oversight by enabling users to correct mistakes, add constraints, or halt undesired trajectories without restarting an interaction. At the same time, streaming inputs can introduce new failure modes: adversarial or accidental late-arriving instructions may steer the model toward policy-violating behavior, and partially generated content may be misinterpreted by users as final or fully-considered. Systems should clearly communicate when outputs are provisional, provide UI affordances for interruption/rollback, and log or surface when late inputs materially change the model's plan.

\textbf{Misuse and dual-use.}
Improved real-time capability can be misused in settings such as persuasive social engineering, automated harassment at scale, or interactive assistance for wrongdoing. While the method itself is training-free and model-agnostic, enabling more responsive agentic behavior can increase the attractiveness of deploying powerful models in high-stakes contexts. Practical safeguards (rate limits, content moderation, audit trails, and conservative defaults for safety mode-switching) are important when integrating this technique into user-facing products.

\textbf{Privacy and data handling.}
Real-time and mid-stream inputs may include sensitive personal information (spoken or typed). Deployments should minimize retention of raw streams, apply encryption in transit and at rest, and provide clear user controls over logging and deletion. If used in voice assistant pipelines, additional privacy risks arise from continuous capture and transcription, which should be addressed at the system level (on-device processing where feasible, explicit recording indicators, and least-privilege access).

\textbf{Environmental and efficiency impacts.}
AsyncReasoning targets \emph{latency} and interactivity rather than reducing total compute in all cases, and may encourage broader use of reasoning-heavy models in real-time applications. Wider adoption could increase overall inference demand; conversely, improved responsiveness may reduce repeated restarts and duplicated computation in interactive sessions.

Overall, we view AsyncReasoning as enabling more natural interactive reasoning systems while introducing new safety, privacy, and misuse considerations that must be addressed with careful prompting, product design, and standard deployment safeguards.

\newpage
\section*{NeurIPS Paper Checklist}

\begin{enumerate}

\item {\bf Claims}
    \item[] Question: Do the main claims made in the abstract and introduction accurately reflect the paper's contributions and scope?
    \item[] Answer: \answerYes{} 
    \item[] Justification: The claims made in the abstract match the experimental results. 

\item {\bf Limitations}
    \item[] Question: Does the paper discuss the limitations of the work performed by the authors?
    \item[] Answer: \answerYes{} 
    \item[] Justification: The paper explicitly discusses the limitations in Appendix~\ref{app:limitations}. 

\item {\bf Theory assumptions and proofs}
    \item[] Question: For each theoretical result, does the paper provide the full set of assumptions and a complete (and correct) proof?
    \item[] Answer: \answerNA{} 
    \item[] Justification: The paper does not include theoretical results. 

    \item {\bf Experimental result reproducibility}
    \item[] Question: Does the paper fully disclose all the information needed to reproduce the main experimental results of the paper to the extent that it affects the main claims and/or conclusions of the paper (regardless of whether the code and data are provided or not)?
    \item[] Answer: \answerYes{} 
    \item[] Justification: The paper fully describes the proposed method, including all key components, enabling independent implementation. Experimental settings and hyperparameters are detailed to support reproducibility of the reported results. 

\item {\bf Open access to data and code}
    \item[] Question: Does the paper provide open access to the data and code, with sufficient instructions to faithfully reproduce the main experimental results, as described in supplemental material?
    \item[] Answer: \answerYes{} 
    \item[] Justification: The attached code allows for the reproduction of the results. 

\item {\bf Experimental setting/details}
    \item[] Question: Does the paper specify all the training and test details (e.g., data splits, hyperparameters, how they were chosen, type of optimizer) necessary to understand the results?
    \item[] Answer: \answerYes{} 
    \item[] Justification: The paper provides all key experimental details, including hyperparameters and selection methodologies, both in the main text and appendix, ensuring reproducibility and clarity. 

\item {\bf Experiment statistical significance}
    \item[] Question: Does the paper report error bars suitably and correctly defined or other appropriate information about the statistical significance of the experiments?
    \item[] Answer: \answerNo{} 
    \item[] Justification: We report statistical significance only for setups where it is standard practice due to low sample sizes (e.g. AIME-2025). For the vast majority of our evaluation setups (10 models, 8 benchmarks), we only report average accuracy to save compute resources. That said, since the sample size is publicly known and the benchmarks use binary scores for each individual sample, the reader can infer bootstrap error bars from the reported accuracy. 

\item {\bf Experiments compute resources}
    \item[] Question: For each experiment, does the paper provide sufficient information on the computer resources (type of compute workers, memory, time of execution) needed to reproduce the experiments?
    \item[] Answer: \answerYes{} 
    \item[] Justification: We report GPU types used in our experiments in Section~\ref{sect:experiments_main} and Appendix~\ref{app:extra_ablation}. However, the total runtime in each setup is hard to estimate due to our preemptible compute infrastructure. It is also difficult since the paper contains experiments across 8 benchmarks and 10 models on 3 GPU types. Instead, we report the total GPU hours required for several typical evaluations, along with the hardware specifications (e.g., GPU type) in Appendix~\ref{app:compute}.  
    
\item {\bf Code of ethics}
    \item[] Question: Does the research conducted in the paper conform, in every respect, with the NeurIPS Code of Ethics \url{https://neurips.cc/public/EthicsGuidelines}?
    \item[] Answer: \answerYes{} 
    \item[] Justification: The paper poses no risk of misuse, does not involve crowdsourcing or research with human subjects, etc. 

\item {\bf Broader impacts}
    \item[] Question: Does the paper discuss both potential positive societal impacts and negative societal impacts of the work performed?
    \item[] Answer: \answerYes{} 
    \item[] Justification: The paper discusses potential societal impact in Appendix~\ref{app:impact_statement}. 
    
\item {\bf Safeguards}
    \item[] Question: Does the paper describe safeguards that have been put in place for responsible release of data or models that have a high risk for misuse (e.g., pre-trained language models, image generators, or scraped datasets)?
    \item[] Answer: \answerNA{} 
    \item[] Justification: The only released dataset artifact is our sharded version of MATH-500, which contains mathematical reasoning problems split into partial-input shards for evaluation. The released data is intended solely for reproducible benchmarking of asynchronous-input reasoning behavior. It does not contain personal, sensitive, proprietary, or safety-critical information, and we do not release any models. 

\item {\bf Licenses for existing assets}
    \item[] Question: Are the creators or original owners of assets (e.g., code, data, models), used in the paper, properly credited and are the license and terms of use explicitly mentioned and properly respected?
    \item[] Answer: \answerYes{} 
    \item[] Justification: The paper properly credits the original creators of all used assets (code, data, models) and explicitly mentions their respective licenses. 

\item {\bf New assets}
    \item[] Question: Are new assets introduced in the paper well documented and is the documentation provided alongside the assets?
    \item[] Answer: \answerYes{} 
    \item[] Justification: The attached source code includes the implementation of our proposed method and the evaluation pipeline, along with detailed instructions for running the experiments to ensure full reproducibility of the results. 

\item {\bf Crowdsourcing and research with human subjects}
    \item[] Question: For crowdsourcing experiments and research with human subjects, does the paper include the full text of instructions given to participants and screenshots, if applicable, as well as details about compensation (if any)? 
    \item[] Answer: \answerNA{} 
    \item[] Justification: The paper does not include any crowdsourcing or research including human subjects. 

\item {\bf Institutional review board (IRB) approvals or equivalent for research with human subjects}
    \item[] Question: Does the paper describe potential risks incurred by study participants, whether such risks were disclosed to the subjects, and whether Institutional Review Board (IRB) approvals (or an equivalent approval/review based on the requirements of your country or institution) were obtained?
    \item[] Answer: \answerNA{} 
    \item[] Justification: The paper does not include any crowdsourcing or research including human subjects. 

\item {\bf Declaration of LLM usage}
    \item[] Question: Does the paper describe the usage of LLMs if it is an important, original, or non-standard component of the core methods in this research? Note that if the LLM is used only for writing, editing, or formatting purposes and does \emph{not} impact the core methodology, scientific rigor, or originality of the research, declaration is not required.
    \item[] Answer: \answerYes{} 
    \item[] Justification: Since our work proposed an LLM inference algorithm, it naturally uses LLMs throughout the experiments. We also used llm-as-a-judge in some of the experiments, which is not a part of the core method. We also used LLM for sample rewriting to create MATH-500 sharded. Additionally, we used Claude Code to create interactive visuals that demonstrate AsyncReasoning in action. 

\end{enumerate}

\end{document}